\documentclass{article}

\usepackage{microtype}
\usepackage{graphicx}
\usepackage{subfigure}
\usepackage{booktabs} 

\usepackage{hyperref}
\usepackage{xcolor}         
\usepackage{multirow}
\usepackage{makecell}

\usepackage[accepted]{icml2025} 

\usepackage{amsmath}
\usepackage{amssymb}
\usepackage{mathtools}
\usepackage{amsthm}

\usepackage[capitalize,noabbrev]{cleveref}

\theoremstyle{plain}

\theoremstyle{definition}

\theoremstyle{remark}

\usepackage[textsize=tiny]{todonotes}

\icmltitlerunning{CogReact: A Reinforced Framework to Model Human Cognitive Reaction Modulated by Dynamic Intervention}

\begin{document}

\twocolumn[
\icmltitle{CogReact: A Reinforced Framework to Model Human Cognitive Reaction Modulated by Dynamic Intervention}



\icmlsetsymbol{equal}{*}

\begin{icmlauthorlist}
\icmlauthor{Songlin Xu}{yyy}
\icmlauthor{Xinyu Zhang}{yyy}

\end{icmlauthorlist}

\icmlaffiliation{yyy}{Department of Electrical and Computer Engineering, University of California San Diego, La Jolla, CA 92093, USA}

\icmlcorrespondingauthor{Songlin Xu}{soxu@ucsd.edu}

\icmlkeywords{Machine Learning, ICML}

\vskip 0.3in
]



\printAffiliationsAndNotice{}  


\begin{abstract}
Using deep neural networks as computational models to simulate cognitive process can provide key insights into human behavioral dynamics. Challenges arise when environments are highly dynamic, obscuring stimulus-behavior relationships. However, the majority of current research focuses on simulating human cognitive behaviors under ideal conditions, neglecting the influence of environmental disturbances. We propose \textbf{\textit{CogReact}}, integrating drift-diffusion with deep reinforcement learning to simulate granular effects of dynamic environmental stimuli on human cognitive process. Quantitatively, it improves cognition modelling by considering temporal effect of environmental stimuli on cognitive process and captures both subject-specific and stimuli-specific behavioural differences. Qualitatively, it captures general trends in human cognitive process under stimuli, better than baselines. Our approach is examined in diverse environmental influences on various cognitive tasks. Overall, it demonstrates a powerful, data-driven methodology to simulate, align with, and understand the vagaries of human cognitive response in dynamic contexts.
\end{abstract}

\section{Introduction}
\label{intro}

Modeling human cognition is a fundamental challenge in understanding human behaviors \cite{jaffe2023modelling}. In particular, modeling the effects of environmental dynamics (e.g., stress \cite{cheng2017evaluation} and feedback \cite{costa2019boostmeup}) on cognitive performance could elucidate behavioral responses to tasks \cite{cheng2017evaluation} and inform the design of feedback mechanisms to augment cognition \cite{costa2019boostmeup}. 
However, prior research \cite{jaffe2023modelling,ma2020neural,peterson2018evaluating,battleday2021convolutional,peterson2021using} predominantly concentrates on modeling human cognition under standard and ideal conditions, often neglecting the nuanced impact of environmental stimuli \cite{do2021simulation}. Alternatively, some studies treat environmental stimuli as a constant presence throughout the cognitive process \cite{bourgin2019cognitive}. 

We propose that a more nuanced modeling approach is imperative, particularly when dealing with dynamic stimuli that can fluctuate over time, contingent upon users' performance. This nuanced approach involves stimuli variation at fine timescales, exerting a continuous influence on human cognitive behaviors.
To illustrate, consider an animated visual stimulus conveying time pressure \cite{slobounov2000neurophysiological}. Such stimuli inform users of the passage of time, evoking sensations of pressure. Representing these stimuli as a binary existence indicator would oversimplify their nuanced effects. Therefore, this paper raises a fundamental question: \textbf{How can we simulate the impact of dynamic environmental stimuli on the regulation of human cognitive behaviors with precision at a fine-grained level?}

We address this question starting by examining how dynamic time pressure stimuli \cite{zur1981effect} influence cognitive performance, particularly within the context of a math arithmetic task—a widely utilized benchmark for evaluating human cognition and logical reasoning \cite{lin2011spatial,judd2021training,daitch2016mapping}. The \textit{dynamism} inherent in time pressure feedback encompasses two primary facets. 
Firstly, the presentation of time pressure can be dynamic, involving the delivery of progressively changing visual frames over time (Fig.~\ref{fig: task and feedback illustration}(a)), thereby instilling a sense of urgency. Secondly, the presence of time pressure may vary dynamically across different trials. Since time pressure stimuli represent a well-established feedback modality to modulate human cognitive performance \cite{cheng2017evaluation,slobounov2000neurophysiological,moore2012time,edland1993judgment,whittaker2016don}, modeling such modulation effect holds the promise to offer valuable insights in understanding human cognition \cite{jaffe2023modelling} and facilitating adaptive intervention design for regulating user performance \cite{costa2019boostmeup}.

\begin{figure*}
\centering
\includegraphics[width=1\linewidth]{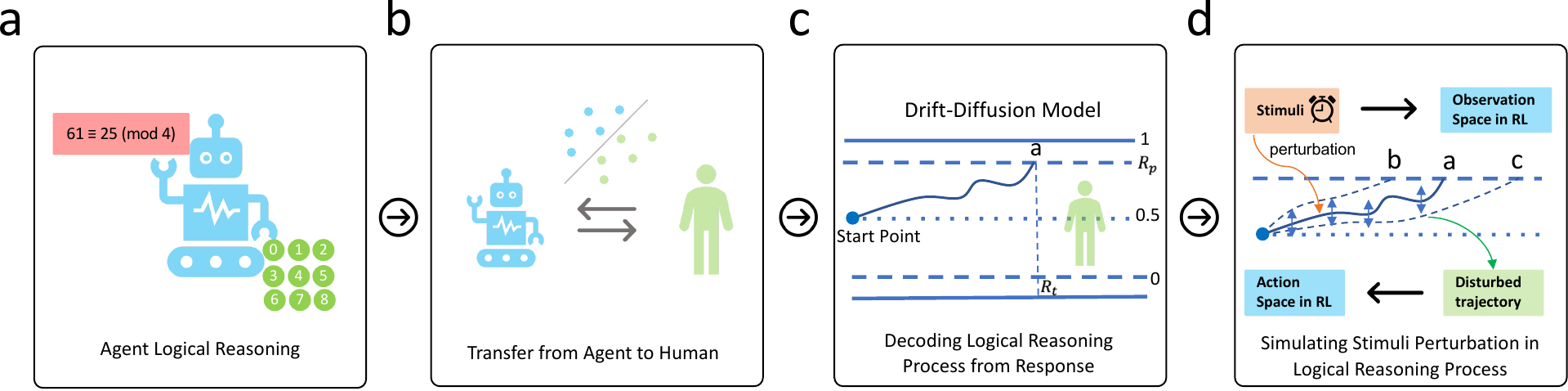}
\caption{\textbf{Illustration of the overall framework}. 
First, we train a logical reasoning agent to solve cognitive tasks without considering users' response. Second, we transfer features extracted from the logical reasoning agent without time pressure to real user choice and response time (initial estimation). Third, the initial estimated response time and predicted choice probability generate evidence accumulation trajectory in the drift-diffusion model. Lastly, the DRL agent simulates influence of stimuli perturbation on cognitive process by taking dynamic environmental stimuli as input and take specific action to modulate evidence accumulation process. When evidence accumulator achieves boundary threshold, the final prediction of response time is generated and DRL agent achieves terminate state.
}
\label{fig: high-level framework}
\end{figure*}

In this paper, we introduce a systematic hybrid framework (\textbf{\textit{CogReact}}) depicted in Fig. \ref{fig: high-level framework}. This framework integrates a classical closed-form cognitive model into a data-driven deep reinforcement learning (DRL) approach, allowing for a comprehensive and explainable simulation of the impacts of dynamic, fine-grained time pressure stimuli.

While neural networks (NNs) are recognized for their proficiency in function approximation and have been applied to model cognitive behaviors \cite{bourgin2019cognitive}, their inherent black-box nature poses challenges in representing the internal mechanisms of the cognitive process. 

To address this limitation, our framework integrates DRL with the drift-diffusion model (DDM), a sequential sampling method widely employed in cognitive modeling \cite{ratcliff2008diffusion, steyvers2019large}. DDM posits that humans make decisions by accumulating evidence until reaching a boundary threshold \cite{fudenberg2020testing}. The simulated choice and response time are then determined based on the corresponding boundary and accumulation time. While DDM excels in representing the cognition process in an explicable and fine-grained manner, it primarily focuses on posterior estimation of user decisions rather than predicting users' future performance under stimuli.

On the other hand, DRL, with NNs at its core, offers a step-by-step interaction environment. This environment enables the incorporation of the fine-grained cognition process inherent in DDM while retaining the function approximation capabilities of NNs. This hybrid approach bridges the gap between the transparency of classical cognitive models and the flexibility of data-driven methods, presenting a promising avenue for modeling the intricate dynamics of cognition under dynamic time pressure stimuli.

In addition, we further show the generalization ability of our framework by extending it into two additional public datasets in more diverse task (decision making, learning) and feedback modalities (numeric, textual). 
In summary, our contribution is three-folded:


\begin{itemize}
    \item We propose \textbf{\textit{CogReact}}, a hybrid framework to incorporate classical cognition models (drift-diffusion model) with deep reinforcement learning to simulate perturbation of environmental stimuli on evidence accumulation process during human cognitive response.
    \item We comprehensively validate the effectiveness of our framework through comparisons with a range of baselines and extensive ablation studies. Additionally, we open-source our codes and newly collected large dataset\footnote{https://github.com/cogreact/CogReact}, comprising 21,157 logical reasoning responses, as a contribution to the research community.
    \item We demonstrate how our framework is adapted to three representative cognitive tasks: mathematical reasoning, decision making, and learning. This adaptation for both discrete user inputs and continuous behaviors establishes a foundation for extending the framework to accommodate diverse user cognitive responses.
    
\end{itemize}

\section{Related Work}
\label{related work}

\textbf{Cognitive Process Models}. 
The existing literature has amassed empirical evidence supporting feasibility of modeling human cognition \cite{de2019overview}. Traditional cognitive models, exemplified by BEAST \cite{erev2017anomalies} and the drift-diffusion model (DDM) \cite{ratcliff2008diffusion, steyvers2019large}, are characterized by closed-form structures. For example, DDM \cite{ratcliff2008diffusion} treats cognitive process as an evidence accumulation process for humans to make decisions, so as to simulate the speed-accuracy tradeoff \cite{heitz2014speed}.

\textbf{Cognitive Simulation with Machine Learning}. 
More recently, there has been a notable shift toward the integration of machine learning techniques \cite{cichy2019deep} for simulating human behaviors \cite{peysakhovich2017using,lake2017building,ma2020neural} across an array of tasks, including visual cognition \cite{cho2023spatially,wenliang2018deep}, categorization \cite{battleday2017modeling}, decision making \cite{binz2024turning,hosoyacognitive,peterson2021using, bourgin2019cognitive}, game strategy \cite{hartford2016deep}, human exploration \cite{binz2022modeling}, word learning \cite{ritter2017cognitive}, etc.

\textbf{Response Time Simulation}. 
Of particular note, recurrent neural networks (RNNs) \cite{jaffe2023modelling,song2017reward, song2016training} have been adapted to execute various cognitive tasks \cite{yang2019task} emulating human performance and the balance between accuracy and response time observed in biological vision \cite{spoerer2020recurrent}.
Recently, \cite{goetschalckx2024computing} computed the human-like reaction time from convolutional RNN using evidential deep learning \cite{sensoy2018evidential}. Furthermore, task-DyVA \cite{jaffe2023modelling} modelled cognitive response time with RNN-based latent dynamical systems. 

These existing models predominantly simulate response time under ideal conditions. In contrast, limited work focuses on modeling the impact of external stimuli perturbations, such as environmental stress, on task performance. We argue that a more nuanced modeling approach is essential, especially when addressing dynamic external stimuli that fluctuate over time based on users' performance. This refined approach requires capturing stimuli variations at fine timescales, which exert a continuous and evolving influence on human cognitive behaviors.

\section{Model and Methodology}
\label{model and method}

\subsection{Math Reasoning Task and Dataset}
\label{subsec: task and dataset}

We used a math arithmetic task with time pressure visual stimuli as our initial model exploration context. The illustration of the task and stimuli is depicted in Fig. \ref{fig: task and feedback illustration} and Appendix. \ref{appendix subsec: dataset}. 
In each math trial, participants were presented with two two-digit numbers and tasked with determining whether their subtraction result was divisible by a given one-digit number. Participants made a binary decision for each trial, with varying settings of time pressure stimuli below.

We collected an extensive dataset encompassing 21,157 valid responses (choice accuracy and response time) from 44 participants engaged in the task (see Fig. \ref{fig:dataset exploration}(a)). To enhance dataset diversity and evaluate our model under dynamic environmental stress, participants were randomly and uniformly distributed across four distinct groups:
\textbf{\textit{None}} Group: Participants experienced no time pressure for any trial.
\textbf{\textit{Static}} Group: Time pressure was consistently applied for each trial.
\textbf{\textit{Random}} Group: There was a 50\% probability of time pressure being applied for each trial.
\textbf{\textit{Rule}} Group: Time pressure was adaptively applied based on users' past performance using a rule-based strategy (more details of such strategy are in Appendix \ref{appendix: Groups}).
This collection has been approved by the Institutional Review Board (IRB) in our local institution. More details are in Appendix \ref{appendix subsec: dataset}.

Our dataset analysis in Appendix \ref{appendix: dataset exploration} revealed that human accuracy remained unaffected by external stimuli, as participants were instructed to prioritize accuracy over speed to control the speed-accuracy tradeoff \cite{heitz2014speed}. Consequently, to model cognitive response due to external stimuli, we focuses on simulating response time rather than choice accuracy, aligning with  \cite{goetschalckx2024computing}.

\subsection{CogReact Framework}
Inspired by exploratory analysis (Appendix \ref{appendix: dataset exploration}) and existing cognitive theories \cite{roseboom2019activity,yang2019task,mickey2014neural}, our framework comprises four key steps, as illustrated in Fig. \ref{fig: high-level framework}. In the initial step, we train a long short-term memory (LSTM)-based logical reasoning agent (termed math agent) to proficiently solve the designated cognitive task. Then the second step involves the knowledge transfer from these trained agents to establish mappings from the LSTM agent to human performance metrics. This results in human response time and accuracy for each trial. Moving to the third step, we employ a fine-grained Drift-Diffusion Model (DDM) to decode human performance, extracting detailed information about response time and accuracy. This step is pivotal in generating the evidence accumulation process (EA) reflective of the underlying cognitive mechanisms. In the final step, we introduce a deep reinforcement learning (DRL) agent to the framework. This agent plays a crucial role in simulating the impact of stimuli perturbation on the evidence accumulation process. By leveraging DRL, we can capture the nuanced dynamics of how external stimuli, such as time pressure, influence the intricate logical reasoning processes modeled by the DDM. We describe details of the first two steps in Section. \ref{subsection: step 1,2}. and the last two steps in Section. \ref{subsection: step 3,4}.

\subsection{Math Agent and Transfer to Humans}
\label{subsection: step 1,2}

To simulate the impact of time pressure, it is imperative to first predict users' baseline performance in ideal conditions without time pressure. Drawing inspiration from prior research that models human subjects' time perception by capturing internal activities in perceptual classification networks \cite{roseboom2019activity}, we have devised a baseline prediction model. Specifically, Roseboom et al. \cite{roseboom2019activity} constructed a neural network functionally akin to human visual processing for image classification. The network was then exposed to input videos of natural scenes, causing changes in network activation. The accumulation of salient changes in activation was subsequently used to estimate duration, effectively gauging the perceived passage of time in the video through a Support Vector Machine (SVM).

\begin{table}
\small
\caption{Evaluation of selected baselines in math reasoning task. For MAPE, we show its mean value (Mean), standard deviation (STD). More complete results with more baselines are in Table. \ref{table:baseline-all}. }
\label{table:baseline}
\setlength{\tabcolsep}{2pt} 
\begin{tabular}{llll}
\toprule
&& \multicolumn{2}{c}{MAPE}\\ 
\cmidrule(r){3-4}
Model Input Type         & Model Type Name    & Mean            & STD \\ 
\midrule
\multirow{4}{*}{\makecell[l]{I. \\ Task: Video, \\ Feedback: Video}}  & hGRU & 0.3335 & 0.2486 \\
 & LSTM + AlexNet & 0.3344 & 0.2602  \\
 & LSTM + VGG-16 & 0.3355 & 0.2708  \\
 & LSTM + ViT-B-16 & 0.3339 & 0.2573 \\
 & MLP + 3D ResNet & 0.3330 & 0.2507  \\
\midrule
\multirow{4}{*}{\makecell[l]{II. \\ Task: Encoded String, \\ Feedback: Video}}  & LSTM-V1 + 3D ResNet & 0.3334 & 0.261 \\
 & LSTM-V2 + 3D ResNet & 0.3376 & 0.2169\\
 & MLP + 3D ResNet & 0.3331 & 0.2550 \\
 & Transformer + 3D ResNet & 0.3306 & 0.2496 \\
 & \textbf{\textcolor{red}{CogReact}} & \textbf{\textcolor{red}{0.2999}} & \textbf{\textcolor{red}{0.2318}}  \\
\midrule
\multirow{4}{*}{\makecell[l]{III. \\ Task: Numeric, \\ Feedback: Video}}  & LSTM-V1 + 3D ResNet & 0.3341 & 0.2617 \\
 & LSTM-V2 + 3D ResNet & 0.3286 & 0.2538 \\
 & MLP + 3D ResNet & 0.3333 & 0.2579  \\
 & Transformer + 3D ResNet & 0.3315 & 0.2526 \\
\midrule
\multirow{4}{*}{\makecell[l]{IV. \\ Task: Numeric, \\ Feedback: Numeric}}  & Decision Tree & 0.3617 & 0.3640\\
 & Linear Regression & 0.3595 & 0.3608 \\
 & LSTM & 0.3059 & 0.2434 \\
 & MLP & 0.3293 & 0.2441 \\
 & Random Forest & 0.3650 & 0.3684\\
 & SVM & 0.3299 & 0.3108 \\
 & Transformer & 0.3052 & 0.2446 \\
  & \textbf{\textcolor{red}{CogReact}} & \textbf{\textcolor{red}{0.2703}} & \textbf{\textcolor{red}{0.2224}} \\
\midrule
\multirow{4}{*}{\makecell[l]{V. \\ Task: Encoded String, \\ Feedback: Numeric}}  & Decision Tree & 0.3639 & 0.3639  \\
 & Linear Regression & 0.3512 & 0.3469\\
 & LSTM & 0.3278 & 0.2478 \\
 & MLP & 0.3333 & 0.2577 \\
 & Random Forest & 0.3600 & 0.3630 \\
 & SVM & 0.3245 & 0.3101 \\
 & Transformer & 0.3299 & 0.2481  \\
\bottomrule
\end{tabular}
\end{table}

Similarly, our baseline prediction model employs an LSTM neural network to address cognitive tasks \cite{yang2019task}. In particular, we train an LSTM-based math answer agent (Fig. \ref{fig: high-level framework}(a)) to learn and respond to math questions, thereby achieving functional similarity with human cognition in math tasks \cite{yang2019task}. The intermediate output of the LSTM layer serves as input features for the SVM, establishing mappings between agents and humans to estimate user choice and response time (Fig. \ref{fig: high-level framework}(b)).
The rationale of this approach is that distinct math questions may pose varying levels of difficulty, leading to user choice biases and variations in response time \cite{hanich2001performance}. The LSTM-based agent has the capacity to capture these potential differences in difficulty levels \cite{mickey2014neural, zaremba2014learning}, 
and the SVM is employed to map these to user choice (via SVC: a classification model in SVM) and response time (via  SVR: a regression model in SVM).
More rationales of the math answer agent and SVM models are in Appendix \ref{appendix: math agent}, \ref{appendix: svm configuration}, Fig.  \ref{fig: detailed hybrid DRL architecture}.

\subsection{Hybrid DRL Agent to Simulate Stimuli Perturbation}
\label{subsection: step 3,4}

To simulate how dynamic time pressure perturbs human logical reasoning process, we conceptualize this process as an evidence accumulation (EA) process in line with the Drift-Diffusion Model (DDM) \cite{ratcliff2008diffusion} (Fig. \ref{fig: high-level framework}(c)). The EA process segments users' cognition into sequential steps, facilitating the fine-grained modeling of dynamic time pressure. The boundary threshold and accumulation time parameters in the DDM are derived from the predicted responses obtained from the previous SVM model.
In order to simulate the dynamic impact of time pressure visual stimuli, we introduce a DRL agent. The visual stimuli are segmented into frames, aligning with the steps in the EA process. For each frame, the specific visual stimuli are applied to the DRL agent (Fig. \ref{fig: high-level framework}(d)), which, akin to how participants' logical reasoning processes may be influenced by each frame of stimuli, modulates the EA process.
In particular, for each frame of time pressure stimuli, the DRL agent adjusts the EA process by introducing positive, neutral, or negative bias (action space of the DRL agent). This modulation may result in the evidence accumulator reaching the boundary threshold either earlier or later. The output from this DRL-modulated EA process serves as the final prediction of user response time (Details in Appendix \ref{appendix: hybrid DRL agent architecture}).

\section{Evaluation}
\label{eval}

\begin{figure*}
\centering
\includegraphics[width=0.8\linewidth]{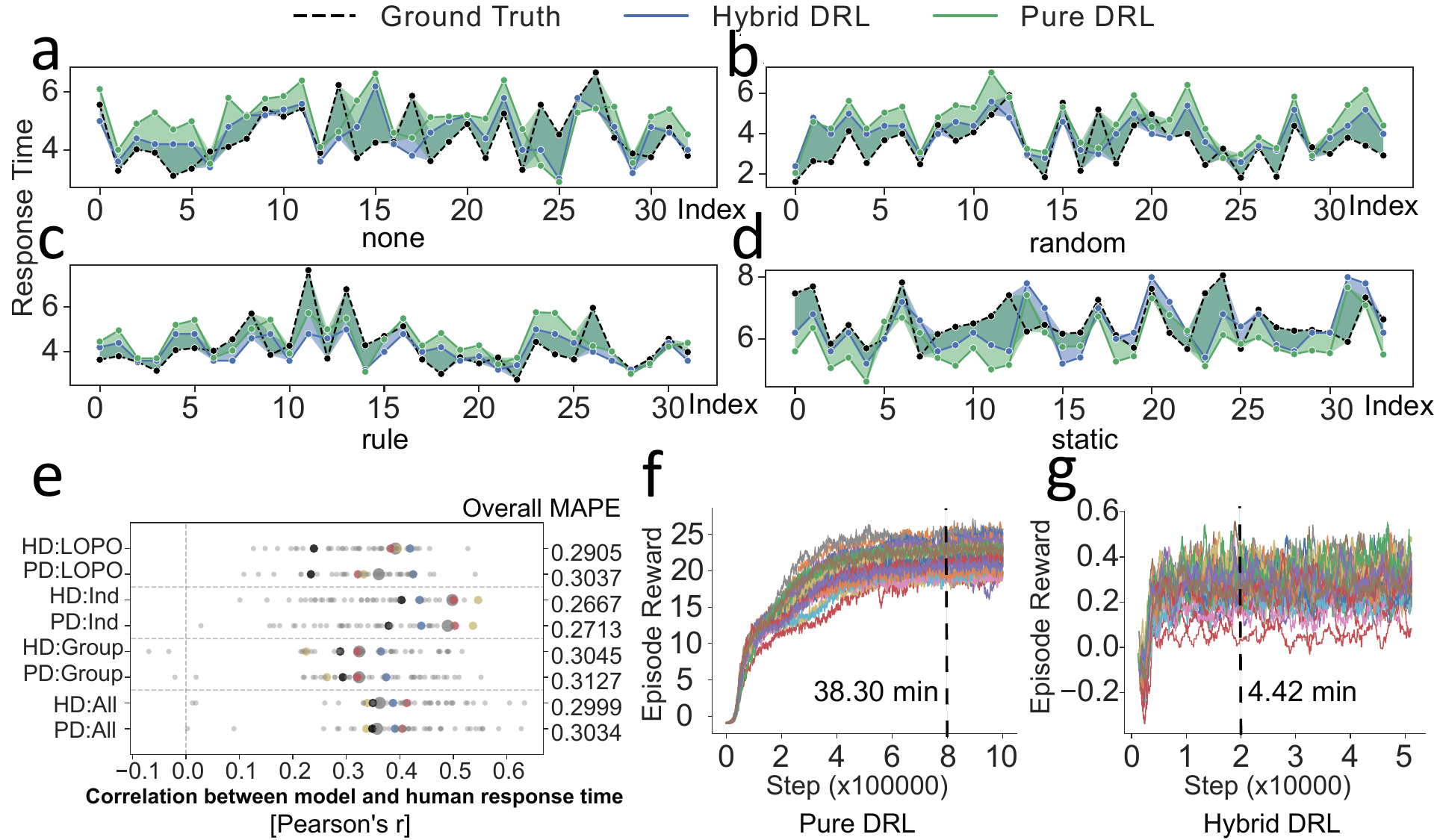}
\caption{
Experimental results in the logical reasoning task.
a,b,c,d: Examples of user response time in chronological order from one participant in each group predicted from Hybrid/ Pure DRL agent in LOPO-level training, compared with ground truth. e: Pearson correlation between predictions from Hybrid/ Pure DRL agent (HD: Hybrid DRL, PD: Pure DRL) and human real response time (ground truth) in four training strategies (All: General-level, Group: Group-level, Ind: Individual-level, LOPO: LOPO-level). Small gray dots, medium dots, and large gray dots represent Pearson correlation of prediction results from each participant's testing set, each group's testing set (red:\textit{none}, yellow:\textit{static}, black:\textit{random}, blue:\textit{rule}) and whole testing set, respectively. The right y axis depicts overall average MAPE of two agents in four training strategies. f,g: Training curve for Pure DRL (f) and Hybrid DRL (g) model. 
}
\label{fig:example curve and train curve}
\end{figure*}

\subsection{Human Response Time Simulation Performance}
We first demonstrate the effectiveness of our \textbf{\textit{CogReact}} framework in human response time simulation by comparing with baselines using different stimuli encoding schemes. 

The model input is composed of three parts: math task stimuli, environmental feedback stimuli (time pressure), and task question ID. The question ID is in numeric value to indicate the trial number for participants in the math task. Our exploratory analysis in Appendix \ref{appendix: dataset exploration} has depicted the relevance of question ID in human response time. 

In \textbf{\textit{CogReact}}, the math task stimuli are represented by one-hot encoded textual strings and the feedback stimuli are represented by videos. However, there are also different ways to extract features from the model input. For example, we can treat both task stimuli and feedback stimuli as numeric value directly or we can put both math task stimuli and feedback stimuli into the whole video as model input, just like what humans watch in the task. Therefore, we traverse five types of model input to represent the features of task stimuli and feedback stimuli and use corresponding baseline models, as depicted in Table. \ref{table:baseline} and Table. \ref{table:baseline-all}. More details of each model input type and the training/testing hyperparameters/process are depicted in Appendix \ref{appendix: baseline models}. 

When encoding both task stimuli and feedback stimuli into a whole video, we use hGRU \cite{goetschalckx2024computing}, LSTM with pre-trained vision models \cite{jaffe2023modelling}, and MLP with pre-trained 3D ResNet \cite{bourgin2019cognitive} as the baseline. These models are adapted into our problem corresponding to the recent State-of-the-Art (SOTA) models in human decision making \cite{bourgin2019cognitive} and response time prediction \cite{goetschalckx2024computing,jaffe2023modelling}. Similarly, for other types of model input, we also use related SOTA models in the specific input type domain. More details of  baseline models and adaptation into our problems are depicted in Appendix \ref{appendix: baseline models}.

We use Mean Average Percentage Error (MAPE) instead of Mean Squared Error (MSE) to evaluate the response time difference between real humans and simulations because human response time comes with high individual differences \cite{faust1999individual}. Therefore, for deep learning models, we use MAPE loss function instead of MSE loss function. Training details are in Appendix \ref{appendix: baseline models}.

Results are depicted in Table. \ref{table:baseline} and Table. \ref{table:baseline-all}, showing that \textbf{\textit{CogReact}} in both Type II and Type IV has the best response time prediction performance (lowest MAPE) by comparing with other models in both the same and different model input types. 
Specifically, CogReact in Type IV has a more efficient representation (numeric encoding) during the model inference process and achieves the best performance.
We also perform statistical analysis in both Kolmogorov-Smirnov test and Permutation test because they do not necessarily assume the data to be normally distributed. We applied them in all baselines. Results in Table. \ref{table:baseline-statistics-string-math} show that CogReact in Type II achieves significantly lower MAPE ($p < 0.001$) than most baselines except for LSTM/Transformer in Type IV (numeric for task/feedback). This is due to different task/feedback modality encoding since CogReact uses Type II (Task: String, Feedback: Video). When we apply CogReact to numeric encoding of Type IV as well, results in Table. \ref{table:baseline-statistics-numeric-math} show that ours can also achieve significantly lower MAPE ($p < 0.001$) than LSTM/Transformer in Type IV. 

\textbf{MAPE Variance}. 
 Despite the superiority, we observed high variance in MAPE, largely driven by individual differences across users and variability in the math trials they completed. These factors interact, amplifying prediction error variance.
To isolate their effects, we held one factor constant while averaging over the other:
(a) Fixing Users: Averaging predicted and actual response times across all trials per user, then computing MAPE, yielded a mean of 0.1388 (STD: 0.0641, 95\% CI: [0.0316, 0.2378]).
(b) Fixing Trials: Averaging across users per trial and computing MAPE resulted in a mean of 0.1403 (STD: 0.0865, 95\% CI: [0.0292, 0.2855]).
Both approaches significantly reduced variance compared to the overall results (Mean: 0.2703, STD: 0.2224, 95\% CI: [0.0093, 0.7631]), suggesting that the high variability stems from the interaction of user and trial differences.

The performance improvement benefits from the whole framework including useful features from the math logical reasoning agent and the integration of the drift-diffusion model in the DRL agent to simulate feedback stimuli in a fine-grained manner. In what follows, we run ablation studies to show the unique importance of each component.

\subsection{Importance of Task Encoding with Math Agent}
To demonstrate that the math answer agent indeed captures representative features from  math questions, in the first ablation study, we compare SVM models (second step in our framework) with two additional settings where the SVM models do not take features captured from the math answer agent as input. Instead, they take raw three-digit numbers from the math questions or one-hot encoded vectors (same as the input of the math logical reasoning agent in Appendix \ref{appendix: math agent}) of raw numbers as input, along with the question id. The SVM performance in the three settings is depicted in Table. \ref{table:svm ablation}. 
Notably, SVM models with features from the math answer agent exhibit significantly higher accuracy (0.9613) and F1-score (0.8996) for user choice prediction and lower MAPE (0.3652) for response time estimation than other settings, underscoring the effectiveness of the math answer agent in capturing representative math question features and the feasibility of predicting user baseline performance in ideal conditions without environmental stimuli with SVM.

\begin{figure*}
\centering
\includegraphics[width=0.9\linewidth]{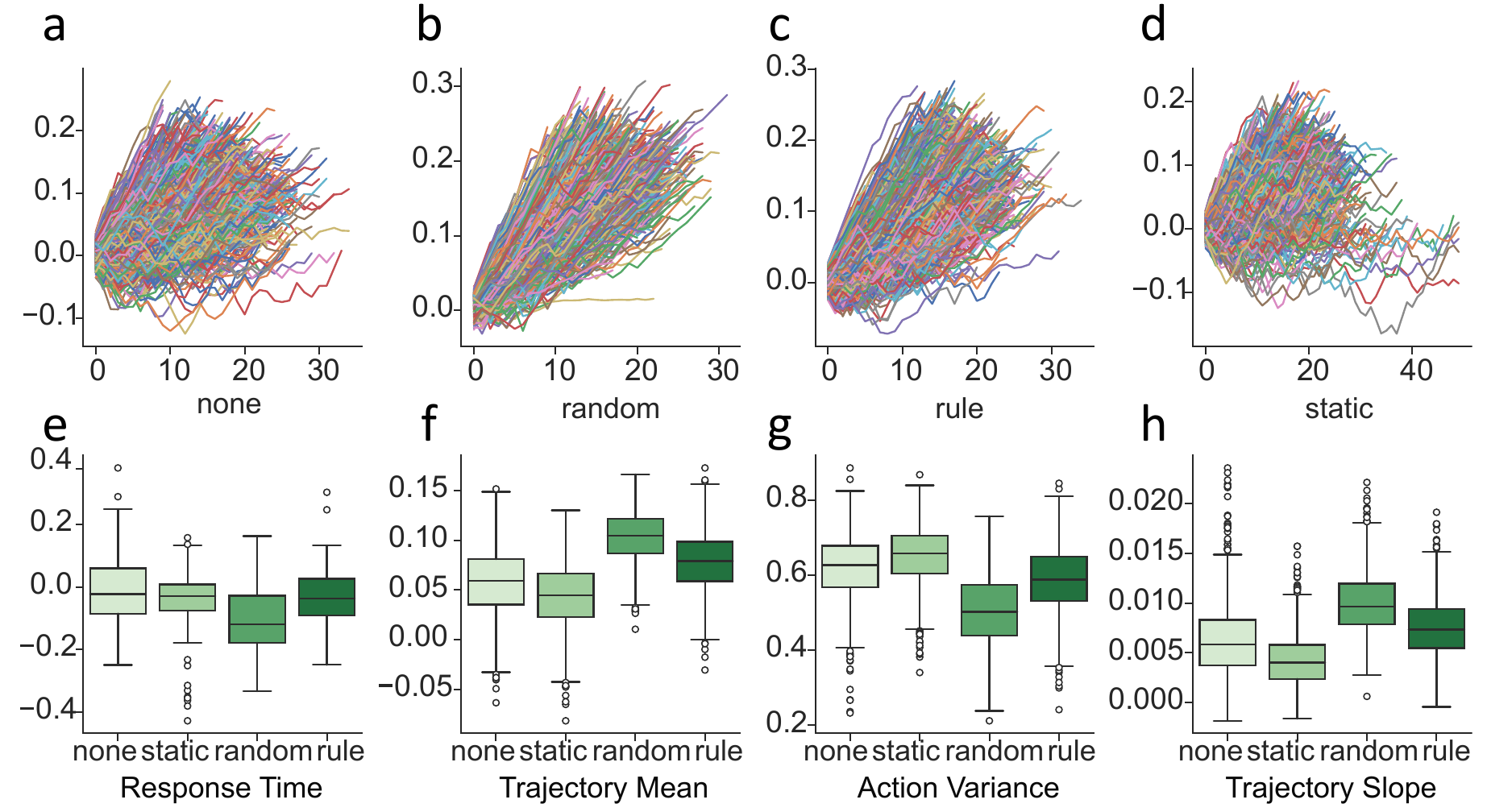}
\caption{
a,b,c,d: Time pressure effect trajectories of four groups, respectively. e: Box plot of relative response time change across four groups in the whole dataset. f,g,h: Box plot of mean value of time pressure effect trajectories (f), standard deviation of action trajectories (g), slope of time pressure effect trajectories (h) of four groups in predicted testing dataset by Hybrid DRL agent. The slope of one trajectory is calculated from the start point to the end point of the trajectory.
}
\label{fig:explanation}
\end{figure*}

\subsection{Why Does the Logical Reasoning Agent Work?}
The second ablation study explores why the math logical reasoning agent could extract useful features from math questions (in the first step of our framework). We answer this question by exploring its math task solving performance under different number of output neurons from LSTM layer.

Note that the math answer agent aims to solve math tasks correctly instead of predicting human choice. In short, given one math question as input, it could directly output the arithmetic reasoning answer. Therefore, its training and testing have no correlation with real users' response. Hence, we prepare a separate dataset that is independent from humans' dataset to train the agent. Finally, we traverse all possible combinations of three numbers in math questions and get a dataset including 20414 samples, which is split into training set (80\%) and testing set (20\%).
Given that the first two numbers of math questions are both two-digits, the arithmetic reasoning result is chosen from 0 to 8. Consequently, the ground truth encompasses 9 classes. 

We experimented with different numbers of output neurons (32, 64, 128, 256) from the LSTM layer. After 100 epochs of training, the logical reasoning agent with 256 neurons achieved remarkable results, attaining a training loss of 0.0001 and 100\% accuracy (Fig.\ref{fig:math agent result}(b)). The confusion matrix (Fig.\ref{fig:math agent result}(a)) for the testing set also demonstrates that this neuron configuration yields over 99\% accuracy for all classes, resulting in an overall test accuracy of 99.93\%. Moreover, even for other neuron number, the test accuracy is also high enough (more than 95\%).
These outcomes affirm that the LSTM-based logical reasoning agent adeptly solves math arithmetic problems in the majority of cases. This aligns with existing work \cite{mickey2014neural, zaremba2014learning}, which demonstrated the capacity of neural networks to learn mathematical equivalence. The success of the logical reasoning agent in solving arithmetic problems lays a foundation and explains its capability for extracting representative features from math questions to construct cognition models.

\subsection{Importance of Integrating DDM into DRL Agents}
\label{sec: ddm ablation}
The third ablation examines the importance of DDM in DRL agents, to simulate the perturbation of external stimuli on human response time in a fine-grained manner. 
We introduce a baseline DRL model called \emph{pure DRL agent}, which does not incorporate the DDM. Specifically, this pure DRL agent does not segment time pressure visual stimuli into frames. Instead, for each trial from the dataset, it directly takes the entire time pressure visual stimuli as input and outputs one action representing the overall change in response time due to time pressure. The final estimation of regulated response time is the sum of this action and the basic response time estimated by the SVR models (details in Appendix \ref{appendix: pure DRL architecture}, Fig. \ref{fig: pure drl model}). Moreover, we also directly remove the whole hybrid DRL agent and only use SVR models to predict response time as another ablation baseline.

We employ both MAPE and Pearson correlation to compare the performance of the hybrid DRL and pure DRL agents. Four model training strategies are used below: (a). \textbf{General-level} involves splitting the entire dataset into training (80\%) and testing (20\%) sets for overall model evaluation. (b). \textbf{Group-level} trains and tests a specific model using data from each group, revealing performance across different time pressure stimuli. (c). \textbf{Individual-level} trains and tests a model using data from a specific participant, assessing personalized model feasibility incorporating subject-specific behavioural differences. 
Shuffling is applied to training and testing sets to prevent overfitting artifacts. (d). As shuffled testing disrupts the temporal trend of user response time across different math trials, we incorporate \textbf{Leave-One-Participant-Out (LOPO)-level} as an additional training strategy. This strategy selects all data from one participant as the testing set and uses data from other participants in the same group as the training set. By traversing every participant's data as the testing set, we ensure a comprehensive assessment of model performance in capturing temporal trends of response time.

Fig. \ref{fig: comparion with pure DRL results} illustrates the average MAPE of the testing set for each individual user (a,b,c,d) and each group (e,f,g,h).  
Both the hybrid DRL and pure DRL agents show improvement in response time estimation compared to SVM results. However, the hybrid DRL agent consistently achieves lower MAPE compared to the pure DRL agent in most cases, indicating the superiority of the hybrid DRL agent in response time estimation.
The overall average MAPE for the entire testing set by both agents is depicted on the right y-axis of Fig. \ref{fig:example curve and train curve}(e), further supporting this conclusion. Fig. \ref{fig:example curve and train curve}(e) also reveals that the hybrid DRL agent exhibits larger Pearson correlation in individual testing sets (small dots), group testing sets (medium dots), and the whole testing set (large dots) compared to the pure DRL agent in most cases, across all four training strategies. Both MAPE and Pearson correlation demonstrate the superior performance of the hybrid DRL agent in modeling the effect of time pressure stimuli.

To compare which agent design better captures the trend of response time change in users' overall tasks, we visualize the prediction results and real user response time for the testing set from one participant of each group in LOPO-level in chronological order (Fig. \ref{fig:example curve and train curve}(a,b,c,d)). The resulting curves clearly demonstrate that the hybrid DRL agent more accurately captures the trend of user response time compared to the pure DRL agent.

\subsection{Training Efficiency}

The training curves for both hybrid and pure agents are presented in Fig. \ref{fig:example curve and train curve}(f,g). The pure DRL and hybrid DRL agents converge at approximately 800,000 steps and 20,000 steps, respectively. It is important to note that the meanings of one step differ between the two agents. For the hybrid DRL agent, one step represents one frame of time pressure stimuli during one trial, whereas one step for the pure DRL agent represents the entire trial. Consequently, a direct comparison of steps is not meaningful. Instead, we compare the training time required for both agents to achieve convergence on the same hardware (GeForce RTX 2080 Ti) and the same dataset.
The results in Fig. \ref{fig:example curve and train curve}(f,g) indicate that the hybrid DRL agent converges in approximately 1/10 of the time compared to the pure DRL agent (4.42 minutes vs. 38.30 minutes). This outcome underscores the advantage of incorporating an explicit cognitive model (i.e., the DDM) in the hybrid DRL agent to improve training efficiency.

\subsection{Interpretability}

An essential advantage of the cognition-inspired hybrid DRL agent is its interpretability, compared with deep learning models and the pure DRL agent, which directly output estimated response time changes for each trial, obscuring the internal mechanism regarding how time pressure stimuli modulate the logical reasoning process. In contrast, the hybrid DRL agent can generate a trajectory of the time pressure effect on response time corresponding to the users' logical reasoning process. Therefore, visualizing the trajectories of the hybrid DRL agent enables the extraction of new insights into how time pressure stimuli affect the human logical reasoning process.

We explore this benefit in Fig. \ref{fig:explanation}(a,b,c,d,e,f,g,h). Here, the \textit{action trajectory} represents the trajectory of actions taken by the hybrid DRL agent during one episode, with each episode corresponding to one math trial of users. The \textit{time pressure effect trajectory} is the accumulated actions multiplied by $\delta_p$. $\delta_p$ represents one unit of evidence per step, transforming the normalized action value into the evidence accumulation process. We visualize the time pressure effect trajectories across the four groups in Fig. \ref{fig:explanation}(a,b,c,d). Each curve represents one trajectory predicted by the hybrid DRL agent during one trial.

We observe that the time pressure effect trajectories are more concentrated in the \textit{random} and \textit{rule} groups but divergent in the \textit{none} and \textit{static} groups (Fig. \ref{fig:explanation}(a,b,c,d)). This suggests that participants in the \textit{random} and \textit{rule} groups, especially the \textit{random} group, are better regulated by the corresponding type of time pressure stimuli, resulting in similar trends in all time pressure effect trajectories in this group. Quantitatively, the \textit{random} group has the lowest standard deviation (STD) of action trajectories (Fig. \ref{fig:explanation}(g)) and the highest average value and slope for the time pressure effect trajectories (Fig. \ref{fig:explanation}(f,h)). These findings in the simulation results indicate that the \textit{random} group experiences the most effective regulation of user cognition performance.

This observation aligns with the expectation that users may quickly adapt to \textit{none} or \textit{static} time pressure, ceasing to be regulated by them after a few trials. However, users may not anticipate the time pressure in \textit{random} group, leading to a more prolonged regulation effect. 
This result in the hybrid DRL simulation is also consistent with real human results in our initial exploratory findings (Appendix \ref{appendix: dataset exploration}, Fig. \ref{fig:dataset exploration}(e)), where participants in \textit{random} group demonstrated a significantly larger reduction in response time, compared with other groups. These experiments affirm the hybrid DRL agent's capability to explain and support observations in the real humans' response time performance.

The comparative analysis between the hybrid and pure DRL agent designs across three key aspects (response time estimation performance, training efficiency, and interpretability) highlights the advantages of the hybrid DRL approach in capturing the nuanced dynamics of time pressure stimuli on user response time in logical reasoning process.


\section{Generalization}
We further evaluate the generalization ability of our model in two additional public datasets: CPC18\footnote{https://cpc-18.com/data/} for decision making and PeerEdu for learning \cite{xu2025peeredu}. Table. \ref{tab:task info} depicts the datasets' properties and summaries diverse tasks and feedback modalities for modeling. Dataset details and our framework adaptation are depicted in Appendix. \ref{appendix: subsec: generalization}.

\begin{table*}[]
\small
\caption{Task/feedback information and dataset properties.}
\label{tab:task info}
\setlength{\tabcolsep}{3.5pt} 
\begin{tabular}{llll|lll|lll}
\hline
\multicolumn{4}{c|}{Task Information}                       & \multicolumn{3}{c|}{Simulation Modality}      & \multicolumn{3}{c}{Dataset Information} \\ 
Task Type & Response Type & User Action & Cognitive Response & Task & Feedback & Stage 1 & Source & Size & User \\ \hline
Math Reasoning  & Active  & Binary     & Response Time      & String  & Visual  & Math  Agent    & Ours     & 21,157      & 50      \\ \hline
Decision Making & Active  & Binary     & Response Time      & Numeric & Numeric & Risk Agent & Public          & 30,489     & 240     \\ \hline
Learning        & Passive & Continuous & Curiosity & Textual & Textual & LLM Agent     & Public          & 12,804      & 300     \\ \hline
\end{tabular}
\end{table*}

\begin{figure}
\centering
\includegraphics[width=1\linewidth]{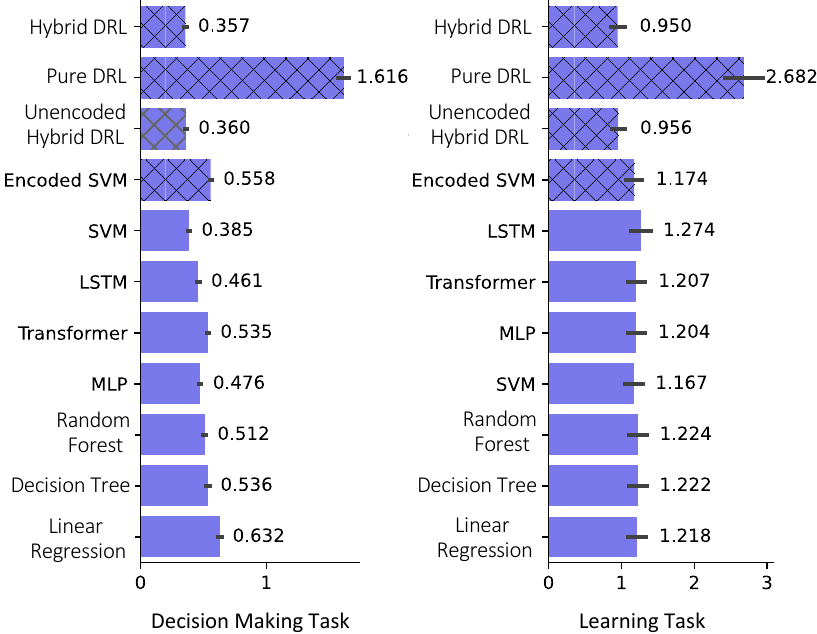}
\caption{Results in decision making(left) and learning(right) task.
}
\label{fig:exp results ext}
\end{figure}

\textbf{Baselines}:
As both datasets lack video input, we adopt the same baselines as Type IV/V in Table \ref{table:baseline}, aligning with baselines commonly used in prior work \cite{bourgin2019cognitive} that benchmarks the CPC18 dataset. For fair comparison, in the PeerEdu dataset, we use the same embeddings from OpenAI's \emph{text-embedding-3-small} model to encode textual data for model input of baselines.

\textbf{Performance}:
Consistent with prior experiments, we use MAPE to evaluate modeling error. As shown in Fig. \ref{fig:exp results ext}, our hybrid DRL model achieves the lowest MAPE across both datasets compared to all baselines, highlighting its effectiveness in generalizing across diverse tasks and feedback modalities. Moreover, the statistical analyses (same with previous evaluations) also show significant improvement of our model over all baseline models in Table. \ref{table:baseline-statistics-peeredu} and Table. \ref{table:baseline-statistics-cpc}.

\textbf{Ablation Study}: 
We further explore the role of each component using variants of our framework, as shown in the textured bar results in Fig. \ref{fig:exp results ext}. The pure DRL variant removes the DDM component from the DRL loop, where the action space directly represents response time/curiosity changes without segmenting the evidence accumulation process (similar to Section \ref{sec: ddm ablation}). The unencoded hybrid DRL variant removes the first step of the framework (risk/LLM agent for embedding extraction), while the encoded SVM variant only includes the first two steps (risk/LLM agent + SVM) without DDM nor the DRL loop.

Fig. \ref{fig:exp results ext} shows that all variants perform worse than the full framework (hybrid DRL), underscoring the unique and critical roles of each component. 
Moreover, the statistical analyses (same with previous evaluations) also show significant improvement of our model over most ablation models in Table. \ref{table:baseline-statistics-peeredu} and Table. \ref{table:baseline-statistics-cpc}.
Removing the first step (unencoded hybrid DRL) causes a slight performance drop, whereas removing the DDM (pure DRL) leads to a significant error increase, performing worse than all baselines. This highlights the dominant role of incorporating DDM into the DRL loop in modeling feedback modulation effects on cognitive responses. Additionally, removing both DDM and DRL (encoded SVM) also degrades performance, emphasizing the importance of the hybrid DRL loop.

Interestingly, the encoded SVM variant (risk/LLM agent + SVM) performs worse than the straightforward SVM baseline (without risk/LLM agent), suggesting potential drawbacks of using risk/LLM embeddings for SVM. However, removing the risk/LLM agent (unencoded hybrid DRL) results in worse performance compared to the full hybrid DRL model. 
This discrepancy suggests that, while the extracted features from the risk/LLM agent may not directly benefit the SVM, they still improve the hybrid DRL model by enhancing and expanding the observation space during the evidence accumulation process.

\textbf{DDM in Deep Learning Models}. Our previous findings demonstrated the advantage of integrating DDM into DRL. To investigate whether this performance gain arises primarily from the deep learning (DL) architecture within DRL or from the reinforcement learning (RL) component itself, we conducted an additional study that directly incorporated DDM into DL models without RL. Specifically, we adapted LSTM, MLP, and Transformer architectures-identical to those used in Baseline Model Type IV-with unchanged hyperparameters, modifying them to predict DDM parameters instead of response times. The final response times were then derived from these predicted parameters. Additionally, we introduced a variant of MLP (MLPv2), which shares the same neural network architecture as our Hybrid DRL model, to assess whether the observed performance gains could be attributed solely to the combination of DL and DDM. 
This experiment was conducted across three datasets, with same statistical analyses performed in previous evaluations. The results (Fig. \ref{fig:result-dl-ddm}) indicate that while DL+DDM integration sometimes outperforms standalone DL models, its performance remains significantly inferior to that of the Hybrid DRL model ($p < 0.001$). These findings highlight that the improvement observed in the hybrid DRL model cannot be attributed solely to DL+DDM integration; rather, the RL component plays a critical and complementary role in achieving superior performance.

\section{Discussion, Limitations and Conclusion}

We propose a computational framework for simulating environmental stimuli perturbations on human cognitive processes, including logical reasoning, decision-making, and learning, across diverse task and feedback modalities. By integrating the drift-diffusion model from cognitive science with deep reinforcement learning, our framework achieves higher simulation accuracy, improved training efficiency, and enhanced interpretability, capturing the granular effects of dynamic stimuli on cognitive processes. The successful adaptation of our framework for continuous behaviors (e.g., curiosity in learning tasks) lays a foundation for extending it to handle continuous user inputs beyond binary responses. This advancement has the potential to offer new insights into machine learning and neuroscience by fostering computational models that better understand human cognition.

One limitation is our focus on response time simulation. Future extensions could incorporate additional cognitive measures, such as user choice, and a wider range of tasks. A potential path involves training task-solving agents, like the logical reasoning agent in math tasks, to emulate human task performance. Building on prior research that highlights the effectiveness of machine learning models in more than 20 cognitive tasks \cite{yang2019task}, our framework could be extended to other domains by linking extracted features from task-solving agents to real user responses using models like SVM. Finally, by dynamically adapting the action and observation spaces to task-specific feedback, the DRL agent could simulate the nuanced effects of stimuli across diverse cognitive scenarios.

\section*{Acknowledgements}

This work is supported by National Science Foundation CNS-2403124,
CNS-2312715, CNS-2128588 and the University of California San
Diego Center for Wireless Communications.

\section*{Impact Statement}

Our framework bridges cognitive science and reinforcement learning by modeling human-like decision-making in dynamic environments. It extends prior work by capturing how environmental stimuli shape cognitive responses, a key challenge in human-in-the-loop AI. This approach enables neuroscientists to simulate cognitive adaptation through reinforcement-based models and supports biologically inspired AI designs and intervention strategies. It also provides cognitive scientists with tools to design behavioral experiments, while helping AI researchers align algorithms with human reasoning. By improving transparency, trust, and usability in adaptive systems, the framework advances human-centered AI and serves as a practical foundation for interdisciplinary research across AI, cognitive science, and neuroscience. 
Future research inspired by this framework could refine its scope to encompass additional cognitive measures, expand its application to diverse task-solving agents, and deepen its alignment with real-world user data. The framework's adaptability to continuous behaviors and dynamic feedback underscores its potential to drive innovation across cognitive modeling, neuroscience, and machine learning. We anticipate that this research will encourage interdisciplinary exploration into cognitive and behavioral modeling, with minimal societal risks. We adhered to appropriate licenses in using public datasets. Our experiment for human data collection in logical reasoning tasks was approved by the Institutional Review Board (IRB) in our local institution. 
The data we model is strictly limited to anonymous behavioral responses under experimental tasks, with no links to real-world identities, nor demographics. As such, the model itself does not encode or have access to demographic-sensitive features, reducing the risk of biased outputs, and also reducing the risk of potential usage for justifying employment, acceptance to college, and so on for targeted people.


\bibliography{0_reference}
\bibliographystyle{icml2025}

\newpage
\appendix
\section{Appendix}

\subsection{Ethics Statement}
Our experiment for human data collection in logical reasoning tasks was approved by the Institutional Review Board (IRB) in our local institution. We do not anticipate any risk during data collection and we have obtained informed consent from all participants beforehand.
Our work may provide insights to integrate classical cognitive theories into machine learning models. In neuroscience, effective computational models for response time could pave the way for understanding many key cognitive behaviors and neurobiological disorders(\cite{goetschalckx2024computing,huys2016computational}). We do not anticipate the negative impact on society in this context.

\subsection{Task and Dataset}
\label{appendix subsec: dataset}

As depicted in Section. \ref{intro}, we used a math arithmetic task with time pressure visual stimuli as our model exploration context. The illustration of the task and stimuli is depicted in Appendix Fig. \ref{fig: task and feedback illustration}. In short, each math trial was composed of two two-digit numbers $Num_1, Num_2$ and one one-digit number $Num_3$, formatted as: $Num_1 \equiv Num_2 \, (\, mod \, Num_3)$. To solve this question, participants first used $Num_1$ to subtract $Num_2$ and judged whether the subtraction result could be divisible by $Num_3$. If it was divisible, they selected "True" button. Otherwise, they selected "False" button. When the time pressure stimuli happened, a progress bar was shown on top of the math question, which added one unit for each second and reset and added again when it accumulated five units. The human response time was then calculated from the time when the math task appeared per trial, to the time when the participants clicked one button to answer it.

We collected an extensive dataset encompassing 21,157 valid responses from 44 participants engaged in the math task (see Fig. \ref{fig:dataset exploration}(a)). To enhance dataset diversity and evaluate our model under dynamic environmental stress, participants were randomly and uniformly distributed across four distinct groups:
\textbf{\textit{None}} Group: Participants experienced no time pressure for any trial.
\textbf{\textit{Static}} Group: Time pressure was consistently applied for each trial.
\textbf{\textit{Random}} Group: There was a 50\% probability of time pressure being applied for each trial.
\textbf{\textit{Rule}} Group: Time pressure was adaptively applied based on users' past performance using a rule-based strategy (more details of such strategy are in Appendix \ref{appendix: Groups}).
Each participant engaged in a two-day study, featuring one exercise session (20 trials) and one formal session (300 trials) per day, when we collected participants' choices and response time per trial. 
This collection has been approved by the Institutional Review Board (IRB) in our local institution. We do not anticipate any risk during data collection and we have obtained informed consent from all participants beforehand.
More dataset details are in Appendix \ref{appendix: dataset info}.

\subsection{Dataset Collection}
\label{appendix: dataset info}

\subsubsection{Participants}
\begin{figure*}
\centering
\includegraphics[width=1\linewidth]{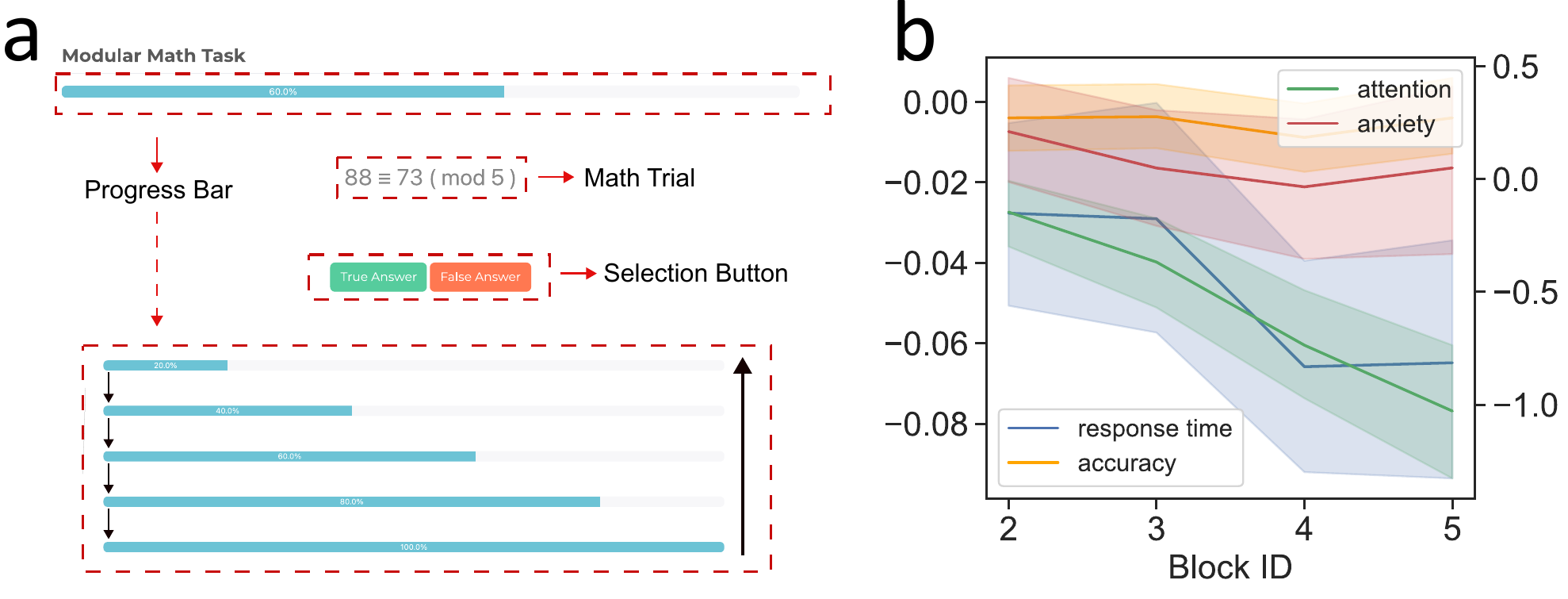}
\caption{a: Math arithmetic task and time pressure feedback. Each math trial is composed of two two-digit numbers $Num_1, Num_2$ and one one-digit numer $Num_3$, formatted as: $Num_1 \equiv Num_2 \, (\, mod \, Num_3)$. To solve this question, participants first use $Num_1$ to subtract $Num_2$ and judge whether the subtraction result could be divisible by $Num_3$. If it is divisible, they select "True" button. Otherwise, they select "False" button. When the time pressure feedback happens, a progress bar will be shown on top of the math question, which adds one unit for each second and reset and add again when it accumulates five units. 
b: Overall trend of relative change of response time/accuracy (left y axis), and attention/anxiety (right y axis), respectively, across 4 blocks.
}
\label{fig: task and feedback illustration}
\end{figure*}

We recruited 50 participants in total (age 21.44 $\pm$ 3.22 y (mean $\pm$ SD); 27 female) from our local institution to finish the math modular task (details in Fig. \ref{fig: task and feedback illustration}(a)). Participants were recruited by email groups in our local institution and came from a variety of majors including engineering, computer science, biology, and so on. Six participants took part in the preliminary study to explore potential configurations of study design, whose results were removed. Other 44 participants were randomly and uniformly divided into 4 groups in order to fully capture the potential effects of time pressure in cognition performance, as described before. Two participants withdrew from the study and three did not finish the study completely. We also removed another three participants' results whose study duration was longer than 3 hours. This was much longer than normal study duration of other participants (within 1 hour) and suggested that participants neither focused on the task nor took this experiment seriously.
Finally, we had 36 participants: \textit{None} group (10), \textit{Static} group (9), \textit{Random} group (7), \textit{Rule} group (10). This study has been approved by the Institutional Review Board (IRB) in our local institution. We have obtained informed consent from all participants before study.

\subsubsection{Procedure}
All participants took part in a two-day study. For each day, they were asked to first finish an exercise session containing 20 math trials and then finish a formal session containing 300 math trials. The exercise session aimed to familiarize the users with tasks and measure users' baseline performance (without time pressure). In the formal session, different time pressure mechanisms were provided for different groups as mentioned above. 
Additionally, participants were requested to rate their current attention/anxiety status on a 7-point Likert scale every 30 trials.
There was also a 5-min rest between exercise session and formal session. 
It took each participant an average of one hour for the study per day.
In the study, participants were told to always take accuracy as the priority and then try their best to answer questions as soon as possible. The compensation rule for each participant (ranging from \$10 to \$100) also prioritized average accuracy over response time in order to encourage participants to follow our instructions. We finally obtained a large data set of 21,157 logical responses after removing invalid user response.

\subsubsection{Math Question Generation and Distribution}
\label{S1}
All math questions are composed of two two-digit numbers ($Num_1$, $Num_2$) and one one-digit number ($Num_3$). We denote the three numbers as $Num_1 = ab$, $Num_2 = cd$, $Num_3 = e$, respectively. So each math question could be denoted as $ab \equiv cd \, (\, mod \, e)$, where $a \in [1,10)$, $b \in [2,10)$, $c \in [1,10)$, $d \in [1,b)$, $e \in [3,10)$. All math questions are randomly generated for each trial. 
We have traversed all possible combinations of math digits in the math question format, which are distributed uniformly in the whole math space for the four groups. 
Participants' accuracy and the provided time pressure feedback are also distributed uniformly.

\subsubsection{Groups}
\label{appendix: Groups}

Here we describe details of four groups in dataset collection. \textit{None} Group: Participants experienced no time pressure for any trial.
\textit{Static} Group: Time pressure was consistently applied for each trial.
\textit{Random} Group: There was a 50\% probability of time pressure being applied for each trial.
\textit{Rule} Group: Time pressure was adaptively applied based on users' past performance using a rule-based strategy. More details about such strategy are depicted below.

Rule-based strategy is designed to provide adaptive time pressure feedback for each trial according to participants' past performance in the \textit{Rule} group. There is a response buffer to update and save user response of most recent 20 trials. For each new user response, it is updated in the response buffer. Then we calculate five metrics (mean response time, delta response time, mean accuracy, push counter, and tolerant counter) in the buffer to decide whether the time pressure feedback is delivered to participants in the next trial. The time pressure feedback only happens if: (a). Mean response time exceeds its threshold RT. Here we use the average response time in exercise session of each specific participant to be RT. (b). Delta response time exceeds its threshold deltaRT = 1 second. (c). Mean accuracy is lower than its threshold accuracy TA. Here we use the average accuracy in exercise session of each specific participant to be TA.
(d). Push counter is lower than its threshold PC = 3. (e). Tolerant counter achieves its threshold TC = 2. 
When the time pressure feedback is decided to be delivered to the participant in the next trial, push counter adds 1 unit and tolerant counter is reset to 0. 

These five metrics aim to ensure that time pressure feedback does not increase user response time but could increase user accuracy. Push counter and tolerant counter are designed to avoid introducing too much distraction to users. The strategy tolerates for a few trials and does not deliver time pressure feedback even if the first three metrics achieve the threshold. After the tolerant counter achieves the TC threshold, it delivers time pressure feedback. In addition, if the strategy delivers time pressure for too many times (exceeding PC threshold), the time pressure feedback is still not delivered to users. Therefore, rule-based strategy is a relatively conservative strategy which cares more about avoiding introducing additional distraction to users.

\begin{figure*}
\centering
\includegraphics[width=1\linewidth]{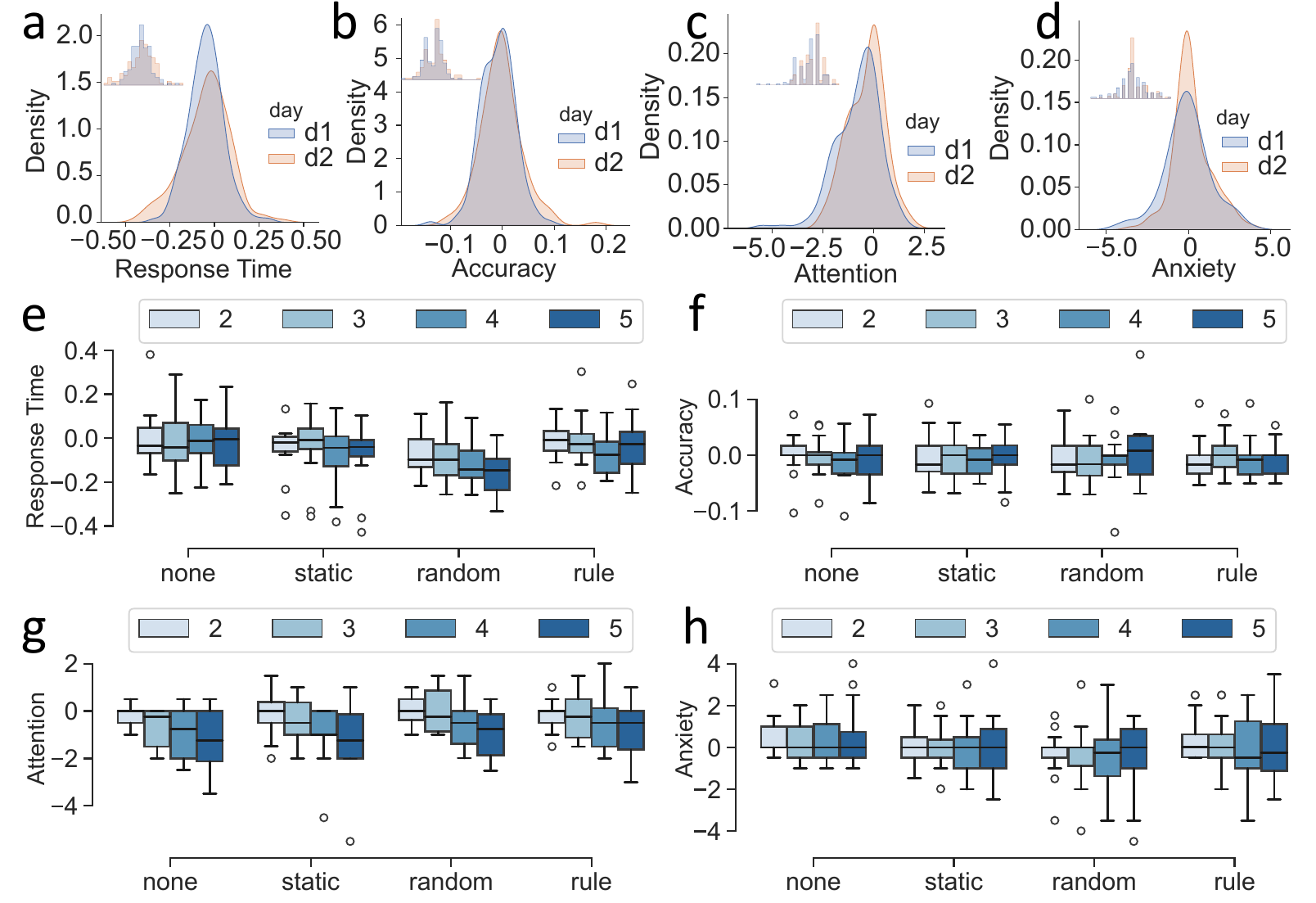}
\caption{
a,b,c,d: overall distribution of relative change of response time (a), accuracy (b), attention (c), and anxiety (d), respectively, across 2 days.
e,f,g,h: box plot of relative change of response time (e), accuracy (f), attention (g), and anxiety (h), respectively, across 4 groups and 4 blocks. 
}
\label{fig:dataset exploration}
\end{figure*}

\subsection{Dataset Exploration}
\label{appendix: dataset exploration}
To investigate the impact of different time pressure stimuli on cognition performance, we conducted an initial exploratory analysis on the dataset. To mitigate the influence of chance factors, we divided the 300 trials of the formal session into five blocks of equal size and calculated the block-wise averages for accuracy, response time, attention, and anxiety scores. Recognizing the inherent variability in users' baseline performance, we aimed to elucidate the impact of time pressure across different groups by comparing the \textit{relative change} in user performance and status across the four groups.
Specifically, let $R_i$ denote the average result of $Block_i$, where $R_1$ ($Block_1$) represents the baseline performance. The final relative result for $Block_i$ ($i>1$) is $(R_i-R_1)/R_1$ for accuracy and response time change and $R_i-R_1$ for attention and anxiety change. This adjustment accounts for the fact that attention/anxiety scores linearly reflect user status, while response time/accuracy changes need to be normalized against participants' individual baseline performances.
The obtained results were then analyzed using repeated-measures ANOVA. To discern specific differences, Bonferroni-corrected paired post hoc t-tests were employed for pairwise comparisons between the groups, enabling a thorough exploration of the impact of different time pressure stimuli on cognition performance and user status.

\subsubsection{Response Time}



In the analysis of between-subjects effects, the ANOVA revealed a significant effect of Group ($F_{3,32} = 3.015, P = 0.044 < 0.05$) (Fig. \ref{fig:dataset exploration}(e)). Specifically, a significant difference was identified between the \textit{none} group (mean $\pm$ SD: $-0.012 \pm 0.021$) and the \textit{random} group ($-0.105 \pm 0.025$) with $p = 0.039 < 0.05$. The \textit{rule} group showed a larger reduction in response time ($-0.034 \pm 0.021$) compared to the \textit{none} group but a smaller reduction compared to the \textit{static} group ($-0.054 \pm 0.022$). Notably, the \textit{random} group exhibited the most substantial reduction in response time. These results suggest that different types of time pressure stimuli may exert varying effects on response time.

Regarding within-subjects tests, a significant effect was observed across blocks ($F_{3,96} = 7.121, P < 0.001$) (Fig. \ref{fig:dataset exploration}(e)), specifically between the following blocks: $Block_2$ ($-0.031 \pm 0.011$) vs. $Block_4$ ($-0.070 \pm 0.014$): $p = 0.023 < 0.05$, $Block_2$ vs. $Block_5$ ($-0.072 \pm 0.014$): $p = 0.026 < 0.05$, $Block_3$ ($-0.033 \pm 0.013$) vs. $Block_4$: $p = 0.008 < 0.01$, $Block_3$ vs. $Block_5$: $p = 0.025 < 0.05$.

No interaction was found between Block and Group ($F_{9,96} = 0.958, P = 0.48$). Furthermore, there was no significant effect of Date ($F_{1,32} = 0.003, P = 0.959$) (Fig. \ref{fig:dataset exploration}(a)), and no other significant interaction effects were identified (all $P > 0.05$). These findings provide valuable insights into the differential impact of time pressure stimuli on response time and underscore the significance of within-subject variations across different blocks.

\subsubsection{Accuracy}


No significant effect was observed in Group ($F_{3,32} = 0.081, P = 0.97 > 0.05$), Block ($F_{3,30} = 0.313, P = 0.816 > 0.05$) (Fig. \ref{fig:dataset exploration}(f)), or Date ($F_{1,32} = 0.861, P = 0.36 > 0.05$) (Fig. \ref{fig:dataset exploration}(b)). Additionally, no other significant interaction effects were identified (all $P > 0.05$). This outcome aligns with expectations, as participants were instructed to prioritize accuracy over response time consistently. Consequently, the accuracy of users' choices should generally be high, while response time may vary depending on the stimuli. The lack of significant effects in these factors supports the study design and participants' adherence to the specified priority in their decision-making process.

The above results suggest that both time pressure stimuli and block number (not experiment date) may impact users' response time. This evidence contributes valuable insights and aligns with
prior theory (\cite{slobounov2000neurophysiological,alexander2003effects}), providing a foundation to inform the design of our cognition model. The observed effects underscore the relevance of considering both math task and question ID in modeling and understanding the dynamics of user response time under varying conditions.

\subsection{Math Logical Reasoning Agent}
\label{appendix: math agent}
Existing work revealed humans' varied performance on different cognitive tasks of diverse difficulty levels (\cite{hanich2001performance}). Therefore, it is essential to first encode features such as difficulty levels of cognitive tasks so that we could model participants' varied responses to different math questions stem from features inherent in the questions. These features may influence user choice and response time even in ideal conditions (i.e., without external stimuli). 
To capture such features, we train a logical reasoning agent capable of solving math questions in a manner similar to humans. Subsequently, feature representations are extracted from the intermediate output of this logical reasoning agent.

\begin{figure*}
\centering
\includegraphics[width=1\linewidth]{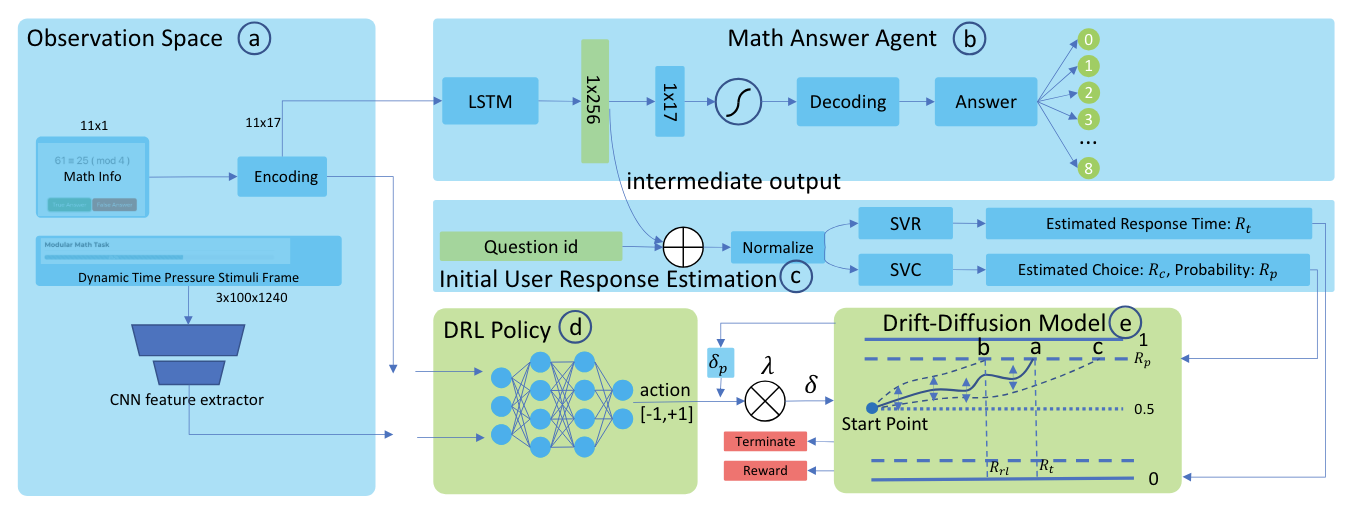}
\caption{The detailed architecture of our CogReact framework. First, we use math questions to train a math answer agent to solve them
without considering users’ response. Second, for each math question, we transfer features extracted from LSTM layer in math
answer agent without time pressure to make predictions of user choice and response time using SVM (initial estimation). The
initial estimated response time and predicted choice probability will generate evidence accumulation trajectory in the
drift-diffusion model. Third, the DRL agent will take math question and each frame of dynamic time pressure stimuli as input
and take specific action to modulate evidence accumulation process. When evidence accumulator achieves boundary threshold,
the final prediction of response time is generated and DRL agent achieves terminate state.
}
\label{fig: detailed hybrid DRL architecture}
\end{figure*}

Illustrated in Fig. \ref{fig: high-level framework} and Fig. \ref{fig: detailed hybrid DRL architecture}, we employ an LSTM-based logical reasoning agent that takes a math question as input and outputs the corresponding answer. For example, given the sequence ``61 $\equiv$ 26(mod 4)'' as input, the agent outputs "3" (the remainder of the subtraction result, "35," of "61" and "26," divided by "4"). It is essential to note the distinction from the data collection process, where users are required to choose whether the subtraction result ("35") of "61" and "26" is divisible by "4"--a binary selection task.

In other words, the logical reasoning agent is trained to answer math arithmetic tasks correctly, rather than to predict user responses. This design choice ensures that the agent learns the potential arithmetic reasoning process and generates representative features of math questions, rather than performing a binary classification task.

The logical reasoning agent is a sequence-to-sequence model based on an LSTM model. Before inputting the math question into the LSTM, the math question is encoded into sequence vectors from original string format. Each math question is denoted as $ab \equiv cd \, (\, mod \, e)$, comprising 11 characters. We use one-hot encoding to deal with the characters. Specifically, each character is mapped into a $1 \times 17$ vector, where the location of this character in a pre-built character dictionary ([`0',`1',`2',`3',`4',`5',`6',`7',`8',`9',`$\equiv$',`(',`m',`o',`d',`)',` ']) is denoted as 1, and other locations are denoted as 0. So we finally obtain the $11 \times 17$ vector for each math question.

For each math question string ($1 \times 11$), we use sequence encoding mentioned above to encode it into a sequence vector ($11 \times 17$), which is then fed into the LSTM model. The hidden unit is 256 neurons, which is then connected with 17 neurons with softmax activation function. Finally, the neuron with the highest probability is the final output answer. We use Keras(\cite{chollet2015keras}) to implement the model (loss function: categorical cross entropy, optimizer: Adam, learning rate: 0.001). 

The logical reasoning agent aims to solve math tasks correctly. In short, given one math question as input, it directly outputs the arithmetic reasoning answer. Therefore, the training and testing of logical reasoning agent have no correlation or connection with real users' response. Hence, we prepare a separate dataset that is independent with users' dataset to train the logical reasoning agent. Finally, we have traversed all possible combinations of three numbers in math questions and gotten a dataset including 20414 samples, which is split into training set (80\%) and testing set (20\%).

\begin{table*}
\caption{Performance of user choice classification of SVC models and response time estimation of SVR models across three math question representations: \textit{Feature} label: SVM (both SVC and SVR) takes features extracted from logical reasoning agent as input, \textit{String} label: SVM (both SVC and SVR) takes encoded vectors of raw math numbers as input, \textit{Digits} label: SVM (both SVC and SVR) takes raw numeric math numbers as input.}
\label{table:svm ablation}
  \centering
\begin{tabular}{lllllllll}
\toprule
& \multicolumn{4}{c}{Choice Classification} & \multicolumn{4}{c}{Response Time Regression (MAPE)}\\ 
\cmidrule(r){2-5}
\cmidrule(r){6-9}
Input & Accuracy & F1-Score & Precision & Recall   & Mean & STD & Lower & Upper \\ 
\midrule
Digits & 0.8107 & 0.0000 & 0.0000 & 0.0000 & 0.3740 & 0.3772 & 0.0121 & 1.4185 \\
String & 0.8174 & 0.0724 & 0.9333 & 0.0377 & 0.3813 & 0.3847 & 0.0135 & 1.4891 \\
\textbf{\textcolor{red}{Feature}} & \textbf{\textcolor{red}{0.9613}} & \textbf{\textcolor{red}{0.8996}} & \textbf{\textcolor{red}{0.8833}} & \textbf{\textcolor{red}{0.9166}} & \textbf{\textcolor{red}{0.3652}} & \textbf{\textcolor{red}{0.3648}} & \textbf{\textcolor{red}{0.0108}} & \textbf{\textcolor{red}{1.3612}} \\
\bottomrule
\end{tabular}
\end{table*}

\subsection{SVM Model Configuration}
\label{appendix: svm configuration}
As previously mentioned, the second step in our simulation framework (Fig.~\ref{fig: high-level framework}) involves transferring features captured by the logical agent to real responses of humans by utilizing SVM models to predict users' baseline performance without time pressure. 
The features comprise the intermediate output of the LSTM layer, with the output neuron number set to 256, resulting in 256 features captured by the math answer agent. During cognition performance analysis, we observed that users' performance is influenced by the block number. Therefore, for each trial, we introduce the question id as an additional input feature, concatenated with the previous 256 features for SVM models. The question id denotes the corresponding trial number in the dataset, resulting in a total of 257 features for predicting user response for each sample/trial. Users' response encompasses both user choice and response time. Consequently, the SVM models consist of a binary SVM classifier (SVC) to predict user choice (True or False selection) and an SVM regressor (SVR) to estimate users' response time.

The SVM model is implemented with scikit-learn(\cite{scikit-learn}). We use default regularization parameter, kernel, and other parameters for both SVM classifier (SVC) and regressor (SVR). The SVR takes 256 features from LSTM layer of math logical reasoning agent as well as question id for input and predicts user response time. The SVC not only predicts user response (choice) but also the probability $R_p$ for each possible response, which serves as the boundary threshold in the drift-diffusion model.

\subsection{Hybrid DRL Agent with Drift-Diffusion Model}
\label{appendix: hybrid DRL agent architecture}

\subsubsection{Drift-Diffusion Model (DDM)}

The DDM assumes that users make decision by accumulating evidence for each choice and make the final selection when the evidence accumulator achieves the threshold. Our framework incorporates the SVM model's predicting results into the DDM. Specifically, we use the output probability of SVC as the accumulated evidence, whose start point is 0.5. The boundary threshold is $R_p$, which is the probability when SVC makes the predictions. Different from traditional DDM that uses Bayesian modelling to draw a distribution of user response time, we need to have a fine-grained trajectory from start point to end point for each math trial to support our reinforcement learning process. Here we use Sigmoid function(\cite{10.5555/646366.689307}) to represent the trajectory from the start point to the end point. 
When users are solving math questions, they are usually more confident given more time to answer (\cite{legg2009math,pajares1994role}). Therefore, we could use a monotonous function to represent the trajectory $T$, i.e. the Sigmoid function. Moreover, we use Brownian motion (\cite{smith2016diffusion}) to add noise into the Sigmoid curve in order to introduce the randomness in decision making trajectory (\cite{smith2016diffusion}). Note that the final simulated trajectory is not always monotonous because such trajectory is modulated and modified by the DRL agent adaptively according to the environmental stimuli.

\subsubsection{DRL Training Loop} 

The DRL training loop is composed of observation space, action space, reward, terminal state, and learning policy. The observation space serves as the model entrance to accept math question information and external stimuli as input. The action space contains a set of potential actions that the DRL agent could take to perform simulation. The reward is used to guide the DRL agent to update its strategy powered by the learning policy to take the optimal action so as to achieve highest possible reward. Terminal state represents the end of one training episode.

\subsubsection{Observation Space}

The observation space consists of two parts: math question information and dynamic time pressure visual stimuli. For each math trial, the math question is encoded as a sequence vector ($11 \times 17$) just like the logical reasoning agent. The dynamic time pressure visual stimuli is segmented into visual frames just like what users perceive in the study. Given frame rate $f$, for each frame $i$, we can obtain the specific image $S_i$ of the visual stimuli for input in the observation space (we set $f=5$). 
In order to encode the frame for input, we use a default CNN feature extractor in Stable Baselines3 (\cite{stable-baselines3}) to extract features automatically from the time pressure image.

\subsubsection{Action Space}

The action space contains one action with continuous numeric value from $-1$ to $1$. The hybrid DRL agent takes one step for each frame $i$. When the output action $a$ is 0, it means that the current time pressure frame has no effect on evidence accumulator in drift-diffusion model. When the output action $a$ is from -1 to 0 or 0 to +1, then it means current time pressure frame leads to negative or positive change $\delta$ on evidence accumulator. The change $\delta$ is obtained from the trajectory of drift diffusion model. Given boundary threshold $R_p$, start point $S_p$, response time $R_t$ and frame rate $f$, the change $\delta$ of evidence accumulation in each frame is $\delta = \lambda \times \delta_p $, $ \delta_p = |R_p-S_p|/(f \times R_t)$, where $\lambda$ is the discounting factor to avoid the DRL agent introducing too aggressive bias.

\subsubsection{Terminal State}

Terminal state happens when the evidence accumulator achieves boundary threshold ($R_t$) or the hybrid DRL agent achieves maximum steps in one episode. Here, one episode represents one math trial in the dataset. Here we set the maximum response time to 10 seconds, consistent with the largest response time in our dataset. So the maximum step number $N = RT_{max} \times f=10 \times 5=50$ steps. If the DRL agent takes $S_n$ steps when the evidence accumulator achieves $R_t$, then the new predicted response time is $R_{rl} = S_n/f$.

\subsubsection{Reward}

For each step during per episode, the hybrid DRL agent only gets reward in the terminal state. For other situations, the reward is 0. The reward mainly aims to encourage the hybrid DRL agent to behave similarly with real users. Therefore, the reward function is: 
\begin{equation}
    r_i = \begin{cases}
    |E_{rl}-E_{svm}|/E_{svm} + P^*, \quad & E_{rl}<E_{svm} \\
    0, \quad & E_{rl} \geq E_{svm}
    \end{cases}
\end{equation}

\noindent where $E_{rl}$ and $E_{svm}$ are the estimated error rate of the hybrid DRL's predicting response time ($R_{rl}$) and the SVM's predicting response time ($R_{svm} = R_t$) compared with real response time ($R_u$) respectively, i.e. $E_{rl} = |R_{rl}-R_u|/R_u$, $E_{svm} = |R_{svm}-R_u|/R_u$. $P^*$ is the penalty caused by terminal state if the hybrid DRL agent's step number exceeds the maximum step threshold ($P^* = -1$). Otherwise, $P^* = 0$.

\subsubsection{Learning Algorithm and Policy}

We use Proximal Policy Optimization (PPO) (\cite{Schulman2017ProximalPO}) as the learning algorithm and multilayer perceptron (MLP) to be the policy for agent training. All hyperparameters and network architectures follow the default settings in Stable Baselines3(\cite{stable-baselines3}). The hybrid DRL model is implemented with PyTorch(\cite{10.5555/3454287.3455008}), Stable Baselines3(\cite{stable-baselines3}), and Gym(\cite{brockman2016openai}). 

\begin{figure*}
\centering
\includegraphics[width=1\linewidth]{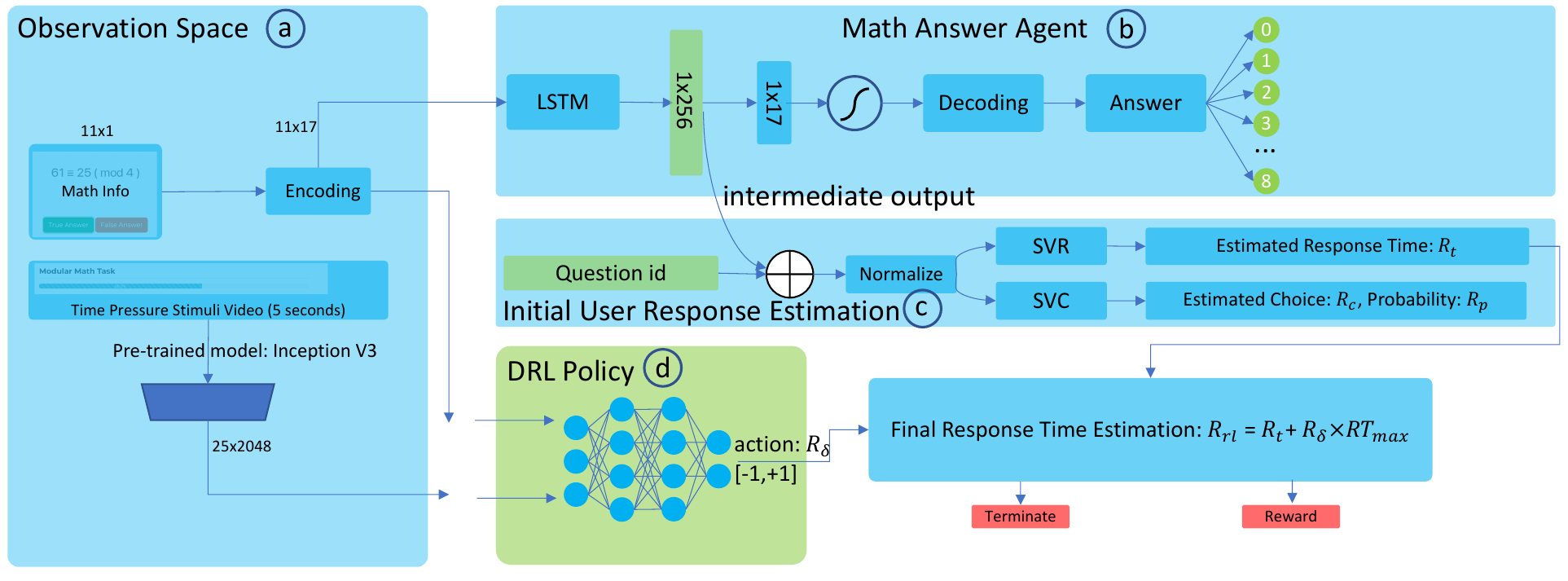}
\caption{The detailed architecture of the pure DRL agent without drift-diffusion model.
}
\label{fig: pure drl model}
\end{figure*}

\subsection{Pure Deep Reinforcement Learning (DRL) Agent}
\label{appendix: pure DRL architecture}
The pure DRL model is implemented with PyTorch(\cite{10.5555/3454287.3455008}), Stable Baselines3(\cite{stable-baselines3}), and Gym(\cite{brockman2016openai}). 

Most parts of the pure DRL agent is the same as the hybrid DRL agent. The main difference lies in the way to represent effect of time pressure in human cognition performance. The hybrid DRL agent segments cognition process of each trial into frames and each action represents specific effect on each frame/step. However, for the pure DRL agent, it directly takes the whole visual stimuli as input and output one action which represents the whole response time change due to time pressure. The final estimation of regulated response time is the sum of this action and basic response time estimated by SVR models.

\subsubsection{DRL Training Loop}
The DRL training loop is similar with the hybrid DRL agent, which is still composed of observation space, action space, reward, terminal state, and learning policy. More details are depicted below.

\subsubsection{Observation Space}
The observation space still consists of two parts: math question information and dynamic time pressure visual stimuli. For each math trial, the math question encoding is the same as the hybrid DRL agent. For time pressure, different from the hybrid DRL agent, the pure DRL agent does not segment visual stimuli into frames. Instead, it takes whole time pressure stimuli video (lasting 5 seconds) as input. We first use a pre-trained Inception-V3 model(\cite{https://doi.org/10.48550/arxiv.1512.00567}) in Keras(\cite{chollet2015keras}) to extract features from this video. The dimension of output features from each frame of the video is $1 \times 2048$. For the whole video, we use the same frame rate as the hybrid DRL agent ($f=5$). So finally we have $5 \ seconds \times 5 = 25$ frames. The final feature dimension of this time pressure visual stimuli in observation space of the pure DRL agent is $25 \times 2048$.

\subsubsection{Action Space}
The action space contains one action ($R_{\delta}$) with continuous numeric value which is normalized into the range from $-1$ to $1$. Different from the hybrid DRL where each step is one frame of user cognition process, here each step of the pure DRL agent is just one trial of users' response. For each trial, user baseline performance is obtained from SVM models. The action of the pure DRL agent represents perturbation for baseline response time ($R_t$) because of time pressure stimuli. Therefore, the final estimation of user response time is $R_{rl} = R_t + R_{\delta} \times RT_{max}$, where $RT_{max} = 10$ is the maximum of user response time in the dataset.

\subsubsection{Terminal State}
The terminal state happens when final estimated response time $R_{rl}$ exceeds normal range (smaller than 0 or larger than $RT_{max} = 10$) or the pure DRL agent achieves maximum steps in one episode. Here, one step represents one math trial in the dataset. Here we set the maximum step number to be 60 steps, which is the same as the trial number of each block in our user study result analysis.

\subsubsection{Reward}
Different from the hybrid DRL agent that could only obtain reward in terminate state, for the pure DRL agent, it gets reward during each step (each trial in user dataset). The reward mainly aims to encourage the pure DRL agent to simulate effect of time pressure visual stimuli that is similar with real users' response. Therefore, the reward function is: 
\begin{equation}
    r_i = \begin{cases}
    |E_{rl}-E_{svm}|/E_{svm} + P^*, \quad & E_{rl}<E_{svm} \\
    0, \quad & E_{rl} \geq E_{svm}
    \end{cases}
\end{equation}

where $E_{rl}$ and $E_{svm}$ are the estimated error rate of the pure DRL's predicting response time ($R_{rl}$) and the SVM's predicting response time ($R_{svm} = R_t$) compared with real response time ($R_u$) respectively, i.e. $E_{rl} = |R_{rl}-R_u|/R_u$, $E_{svm} = |R_{svm}-R_u|/R_u$. $P^*$ is the penalty caused by terminal state if the pure DRL agent's estimated response time exceeds the normal range (0 to 10 seconds) ($P^* = -1$). Otherwise, $P^* = 0$.

\subsubsection{Learning Algorithm and Policy}

We use Proximal Policy Optimization (PPO)(\cite{Schulman2017ProximalPO}) as the learning algorithm and multilayer perceptron (MLP) to be the policy for agent training. All hyperparameters and network architectures follow default settings in Stable Baselines3(\cite{stable-baselines3}).

\subsection{Baseline Models}
\label{appendix: baseline models}

Our baseline models are adapted into our problem corresponding to the recent State-of-the-Art (SOTA) computational models in human decision making (\cite{bourgin2019cognitive}) and response time prediction (\cite{goetschalckx2024computing,jaffe2023modelling}).

The whole dataset is first split into raw training (80\%) and test set (20\%). The raw training set is then split into model training set (80\%) and validation set (20\%). The validation set is used to select the best epoch.

All neural network-based models use MAPE loss function, Adam optimizer (\cite{kingma2014adam}) with learning rate of 0.001 and batch size of 16. All models are trained on 2 Nvidia RTX A6000 GPUs (48GB GPU memory). All neural network models are implemented by PyTorch (\cite{10.5555/3454287.3455008}) and other machine learning models are implemented by scikit-learn (\cite{scikit-learn}).

\begin{itemize}
    \item Baseline Type 1: Model Input Format: Task: Video, Feedback: Video, Question ID: Numeric Value. 
    \begin{itemize}
        \item hGRU (\cite{linsley2018learning}): This model comes from (\cite{goetschalckx2024computing}) to simulate human response time in visual tasks. We use (\cite{goetschalckx2024computing,linsley2018learning}) to implement this model. The original hGRU model accepts image as input. We adjust the dimensions to accept video (including both task and time pressure visual feedback) as input. This model is trained from beginning without pre-trained models. The output of hGRU model is then concatenated with question ID for input into a linear layer (64 neurons) to predict response time. Each epoch takes about 40 minutes for training. We report the results for the best epoch out of 30 (based on performance on the validation set). 
        \item LSTM + AlexNet: This model is based on (\cite{jaffe2023modelling}) that uses LSTM to simulate human response time in cognitive tasks. Here we use the same LSTM configurations as (\cite{jaffe2023modelling}). To adapt it to accept video as input, we first use pre-trained AlexNet (\cite{krizhevsky2012imagenet}) from TorchVision (\cite{10.5555/3454287.3455008}) to extract features from each frame of the video. The sequence of features from all frames are then input into LSTM layer. The output of the LSTM layer is then concatenated with question ID for input into a linear layer (64 neurons) to predict response time. Each epoch takes about 40 minutes for training. We report the results for the best epoch out of 30 (based on performance on the validation set). 
        \item LSTM + VGG-16: This model is similar with LSTM + AlexNet but we replace the AlexNet with pre-trained VGG-16 (\cite{simonyan2014very}) in TorchVision (\cite{10.5555/3454287.3455008}) to extract visual features from video frames. Each epoch takes about 40 minutes for training. We report the results for the best epoch out of 30 (based on performance on the validation set). 
        \item LSTM + ViT-B-16: This model is similar with LSTM + AlexNet but we replace the AlexNet with pre-trained ViT-B-16 (\cite{dosovitskiy2020image}) in TorchVision (\cite{10.5555/3454287.3455008}) to extract visual features from video frames. Each epoch takes about 60 minutes for training. We report the results for the best epoch out of 30 (based on performance on the validation set).
        \item MLP + 3D ResNet: This model is based on (\cite{bourgin2019cognitive}) that uses MLP to predict human decision making. We follow the same MLP architecture as (\cite{bourgin2019cognitive}). To adapt it to accept video input, we first use pre-trained 3D ResNet (\cite{tran2018closer}) in TorchVision (\cite{10.5555/3454287.3455008}) to extract features from the video directly (instead of each video frame). The extracted features are then concatenated with question ID for input into the MLP model. Each epoch takes about 25 minutes for training. We report the results for the best epoch out of 30 (based on performance on the validation set). 
    \end{itemize}
    \item Baseline Type 2: Model Input Format: Task: Encoded String, Feedback: Video, Question ID: Numeric Value
    \begin{itemize}
        \item LSTM-V1 + 3D ResNet: This model is based on (\cite{jaffe2023modelling}) that uses LSTM to simulate human response time in cognitive tasks. Here we use the same LSTM configurations as (\cite{jaffe2023modelling}). To adapt it to accept video input, we first use pre-trained 3D ResNet (\cite{tran2018closer}) in TorchVision (\cite{10.5555/3454287.3455008}) to extract features from the video directly (instead of each video frame). The extracted feedback video features are then concatenated with both math task string with one-hot encoding and question ID for input into the LSTM model. The output of the LSTM layer is then passed into a linear layer (64 neurons) to predict response time. Each epoch takes about 12 minutes for training. We report the results for the best epoch out of 30 (based on performance on the validation set). 
        \item LSTM-V2 + 3D ResNet: This model is similar with LSTM-V1 + 3D ResNet. The difference is that the extracted feedback video features are first fed into the LSTM layer and then the output is concatenated with both math task string with one-hot encoding and question ID to predict response time. Each epoch takes about 12 minutes for training. We report the results for the best epoch out of 30 (based on performance on the validation set). 
        \item MLP + 3D ResNet: This model is based on (\cite{bourgin2019cognitive}) that uses MLP to predict human decision making. We follow the same MLP architecture as (\cite{bourgin2019cognitive}). To adapt it to accept video input, we first use pre-trained 3D ResNet (\cite{tran2018closer}) in TorchVision (\cite{10.5555/3454287.3455008}) to extract features from the video directly (instead of each video frame). The extracted features are then concatenated with both math task string with one-hot encoding and question ID for input into the MLP model. Each epoch takes about 15 minutes for training. We report the results for the best epoch out of 30 (based on performance on the validation set). 
        \item Transformer + 3D ResNet: This model is similar with MLP + 3D ResNet. The difference is that we replace the MLP model with the transformer model. We follow the default architecture of transformer in (\cite{vaswani2017attention}). Each epoch takes about 12 minutes for training. We report the results for the best epoch out of 30 (based on performance on the validation set). 
    \end{itemize}
    \item Baseline Type 3: Model Input Format: Task: Numeric Value, Feedback: Video, Question ID: Numeric Value. 
    \begin{itemize}
        \item LSTM-V1 + 3D ResNet: This model is based on (\cite{jaffe2023modelling}) that uses LSTM to simulate human response time in cognitive tasks. Here we use the same LSTM configurations as (\cite{jaffe2023modelling}). To adapt it to accept video input, we first use pre-trained 3D ResNet (\cite{tran2018closer}) in TorchVision (\cite{10.5555/3454287.3455008}) to extract features from the video directly (instead of each video frame). The extracted feedback video features are then concatenated with both math task digits and question ID for input into the LSTM model. The output of the LSTM layer is then passed into a linear layer (64 neurons) to predict response time. Each epoch takes about 12 minutes for training. We report the results for the best epoch out of 30 (based on performance on the validation set). 
        \item LSTM-V2 + 3D ResNet: This model is similar with LSTM-V1 + 3D ResNet. The difference is that the extracted feedback video features are first fed into the LSTM layer and then the output is concatenated with both math task digits and question ID to predict response time. Each epoch takes about 12 minutes for training. We report the results for the best epoch out of 30 (based on performance on the validation set). 
        \item MLP + 3D ResNet: This model is based on (\cite{bourgin2019cognitive}) that uses MLP to predict human decision making. We follow the same MLP architecture as (\cite{bourgin2019cognitive}). To adapt it to accept video input, we first use pre-trained 3D ResNet (\cite{tran2018closer}) in TorchVision (\cite{10.5555/3454287.3455008}) to extract features from the video directly (instead of each video frame). The extracted features are then concatenated with both math task digits and question ID for input into the MLP model. Each epoch takes about 15 minutes for training. We report the results for the best epoch out of 30 (based on performance on the validation set). 
        \item Transformer + 3D ResNet: This model is similar with MLP + 3D ResNet. The difference is that we replace the MLP model with the transformer model. We follow the default architecture of transformer in (\cite{vaswani2017attention}). Each epoch takes about 12 minutes for training. We report the results for the best epoch out of 30 (based on performance on the validation set). 
    \end{itemize}
    \item Baseline Type 4: Model Input Format: Task: Numeric Value, Feedback: Numeric Value, Question ID: Numeric Value. For this baseline type, all input features (task, feedback, question ID) are directly concatenated into 1D array for input into models. The baseline models in this type are mainly based on (\cite{bourgin2019cognitive}), which presents several machine learning models to predict human decision making with similar model input.
    \begin{itemize}
        \item Decision Tree: We use scikit-learn (\cite{scikit-learn}) to implement this model and follow all default settings in scikit-learn. The training process takes within 10 minutes.
        \item Linear Regression: We use scikit-learn (\cite{scikit-learn}) to implement this model and follow all default settings in scikit-learn. The training process takes within 10 minutes.
        \item LSTM: This model is based on (\cite{jaffe2023modelling}) that uses LSTM to simulate human response time in cognitive tasks. Here we use the same LSTM configurations as (\cite{jaffe2023modelling}). Each epoch takes about 2 minutes for training. We report the results for the best epoch out of 100 (based on performance on the validation set). 
        \item MLP: This model is based on (\cite{bourgin2019cognitive}) that uses MLP to predict human decision making. We follow the same MLP architecture as (\cite{bourgin2019cognitive}). Each epoch takes about 2 minutes for training. We report the results for the best epoch out of 100 (based on performance on the validation set). 
        \item Random Forest: We use scikit-learn (\cite{scikit-learn}) to implement this model and follow all default settings in scikit-learn. The training process takes within 10 minutes.
        \item SVM: We use scikit-learn (\cite{scikit-learn}) to implement this model and follow all default settings in scikit-learn. The training process takes within 10 minutes.
        \item Transformer: We follow the default architecture of transformer in (\cite{vaswani2017attention}). Each epoch takes about 2 minutes for training. We report the results for the best epoch out of 100 (based on performance on the validation set). 
    \end{itemize}
    \item Baseline Type 5: Model Input Format: Task: Encoded String, Feedback: Numeric Value, Question ID: Numeric Value. For this baseline type, the math task questions come with textual string format and get encoded with one-hot encoding (\cite{rodriguez2018beyond}), which are then concatenated with feedback and question ID for input into models. The baseline models in this type are mainly based on (\cite{bourgin2019cognitive}), which presents several machine learning models to predict human decision making with similar model input. The settings of these baseline models (Decision Tree, Linear Regression, LSTM, MLP, Random Forest, SVM, Transformer) are the same as Baseline Type 4.
    
\end{itemize}


\subsection{Further Discussion}
\label{appendix: further discussion}
Modelling dynamics in human cognitive responses to external stimuli is fundamental to understand how the brain dynamically reacts to the environment.
However, the prevailing trend in contemporary research (\cite{jaffe2023modelling,peysakhovich2017using,lake2017building,ma2020neural,mehrer2020individual,golan2020controversial,kumbhar2020anytime,battleday2017modeling,battleday2020capturing,singh2020end,peterson2018evaluating,battleday2021convolutional,peterson2021using, noti2016behavior, bourgin2019cognitive, plonsky2017psychological}) predominantly centers on the modeling of human cognition within standardized and idealized contexts, thereby often neglecting the nuanced influence exerted by external stimuli. Conversely, certain investigations adopt an oversimplified perspective by treating external stimuli as a persistent and unchanging factor throughout the cognitive processes (\cite{bourgin2019cognitive}). A more sophisticated modeling methodology is deemed essential, particularly when addressing dynamic environmental stimuli that exhibit temporal fluctuations contingent upon user performance. This refined approach advocates for a nuanced consideration of stimuli variation at fine temporal scales, thereby perpetuating a continuous impact on human cognitive behaviors.

Our hybrid modeling approach, characterized by the incorporation of Deep Reinforcement Learning (DRL) to emulate external stimuli within the explainable drift-diffusion model at a granular level, takes into account subject-specific and stimuli-specific behavioral distinctions. This distinctive feature sets our framework apart from antecedent studies, which predominantly concentrated on the coarse-grained posterior estimation of decision-making through reinforcement learning (\cite{viejo2015modeling, pedersen2017drift}). The elucidative nature of our framework significantly augments our capacity to comprehend and interpret the intricate interplay between environmental stimuli and cognitive behaviors.

The principles underlying CogReact may be extended to the analysis of neural and physiological responses to external stimuli. Although such data—whether derived from neural activity or wearable sensors—pose significant challenges for direct modeling due to their high-frequency, noisy time-series nature, they hold considerable potential. Specifically, these recordings can serve as proxies for the trajectory of evidence accumulation in human decision-making and other cognitive processes. Mapping these novel forms of evidence accumulation within specific cognitive tasks offers a promising avenue for capturing human cognition at a highly fine-grained temporal and representational level.

\subsection{Generalization in New Tasks and Datasets}
\label{appendix: subsec: generalization}

\subsubsection{Datasets}

We further evaluate the generalization ability of our model in two additional public datasets. The first is CPC18, a widely used benchmark for modeling human cognition in a decision making task \cite{bourgin2019cognitive}. In this dataset, participants engaged in a gambling game where they made binary decisions in each trial. Each choice offers different reward/loss with certain probabilities, and feedback indicates the alternative reward or loss if the participant had selected the other option. Consistent with previous experiments, the model predicts user response times per trial based on the model input including task information (reward/loss probability configurations of two choices) and feedback information (alternative reward/loss), represented as numeric arrays. We totally obtained 30,489 trials from 240 participants with valid response time from the raw dataset.

The second public dataset, PeerEdu \cite{xu2025peeredu}, captures students' cognitive states during learning with external peer feedback. Students watched online video lectures while their cognitive states, were passively recorded using sensors. Specifically, we use one cognitive state named curiosity (continuous value from 0 to 1) for evaluation, which is the most significantly impacted by peer feedback from \cite{xu2025peeredu}. Unlike previous tasks requiring active but binary human choice input, PeerEdu focuses on passive yet continuous cognitive states without students' active input. Peer feedback was delivered by highlighting specific regions on video lecture slides that peer students focused on, updated continuously based on lecture progress.

To simulate curiosity in this learning task, each lecture was divided into transcripts, with each transcript representing a sentence spoken by the lecturer. Task information includes transcript content, while feedback information comprises the highlighted text in peer feedback regions on the slides. The model predicts curiosity for each transcript by taking both task and feedback information (textual format) as input. We totally obtained 12,804 samples from 300 students in PeerEdu, where each sample corresponds to the curiosity of one student during one transcript with specific feedback.

\subsubsection{Framework Adaptation}
To adapt our framework for the decision making task, we replace the math agent with an LSTM-based risk agent in the first step in Fig. \ref{fig: high-level framework}. This risk agent, using the same architecture as before, predicts potential reward/loss (feedback information) from task inputs in each gambling trial, extracting risk features for the SVM model in the second step. The second step keeps the same as Fig. \ref{fig: high-level framework} to predict user choice and response time without feedback, which are used to generate the evidence accumulation process in the DDM step. The DRL loop incorporates an adjusted observation space to handle the task and feedback information in decision making, while maintaining the same action space and reward functions as the previous logical reasoning task for simulating response time changes.

For the learning task adaptation, we replace the math agent in our framework with a large language model (LLM), referred to as the LLM agent, to extract features from textual task and feedback information, leveraging LLMs' strong textual data mining capabilities \cite{wang2023improving}. Specifically, we use OpenAI's \emph{text-embedding-3-small} model to generate embeddings from both task and feedback information, which are then input into SVM models in the second step of Fig. \ref{fig: high-level framework}.
To handle the absence of binary user input, we categorize curiosity values into high or low levels based on their position relative to the median, enabling SVM models to predict curiosity level (SVC: high or low) and actual value (SVR), similar to predicting binary choice and response time in previous tasks. This approach enables adaptation to continuous response modeling without significant changes to the framework and can be extended to other continuous behaviors in the future. The DRL loop incorporates an updated observation space to process embeddings from task and feedback information, with action space and reward functions adjusted to align with the curiosity scale.

\begin{table*}
\caption{Results for all baseline model performance on response time simulation in Math Task. For MAPE, we show its mean value (Mean), standard deviation (STD), 2.5th (Lower) and 97.5th (Upper) percentiles of the MAPE distribution (95\% confidence interval). }
\label{table:baseline-all}
  \centering
\begin{tabular}{llllll}
\toprule
&& \multicolumn{4}{c}{MAPE}\\ 
\cmidrule(r){3-6}
Model Input Type         & Model Type Name    & Mean            & STD & Lower & Upper \\ 
\midrule
\multirow{4}{*}{\makecell[l]{Task: Video \\ Feedback: Video}}  & hGRU & 0.3335 & 0.2486 & 0.0153 & 0.9406 \\
 & LSTM + AlexNet & 0.3344 & 0.2602 & 0.0132 & 0.9954 \\
 & LSTM + VGG-16 & 0.3355 & 0.2708 & 0.0128 & 1.0393 \\
 & LSTM + ViT-B-16 & 0.3339 & 0.2573 & 0.0145 & 0.9852 \\
 & MLP + 3D ResNet & 0.3330 & 0.2507 & 0.0121 & 0.9390 \\
\midrule
\multirow{4}{*}{\makecell[l]{Task: Encoded String \\ Feedback: Video}}  & LSTM-V1 + 3D ResNet & 0.3334 & 0.261 & 0.0151 & 0.9866 \\
 & LSTM-V2 + 3D ResNet & 0.3376 & 0.2169 & 0.0185 & 0.7618 \\
 & MLP + 3D ResNet & 0.3331 & 0.2550 & 0.0125 & 0.9601 \\
 & Transformer + 3D ResNet & 0.3306 & 0.2496 & 0.0145 & 0.9462 \\
 & \textbf{\textcolor{red}{CogReact}} & \textbf{\textcolor{red}{0.2999}} & \textbf{\textcolor{red}{0.2318}} & \textbf{\textcolor{red}{0.0131}} & \textbf{\textcolor{red}{0.8029}} \\
\midrule
\multirow{4}{*}{\makecell[l]{Task: Numeric Value \\ Feedback: Video}}  & LSTM-V1 + 3D ResNet & 0.3341 & 0.2617 & 0.0152 & 0.9923 \\
 & LSTM-V2 + 3D ResNet & 0.3286 & 0.2538 & 0.0147 & 0.9707 \\
 & MLP + 3D ResNet & 0.3333 & 0.2579 & 0.0147 & 0.9731 \\
 & Transformer + 3D ResNet & 0.3315 & 0.2526 & 0.0152 & 0.9615 \\
\midrule
\multirow{4}{*}{\makecell[l]{Task: Numeric Value \\ Feedback: Numeric Value}}  & Decision Tree & 0.3617 & 0.364 & 0.015 & 1.3729 \\
 & Linear Regression & 0.3595 & 0.3608 & 0.0113 & 1.3399 \\
 & LSTM & 0.3059 & 0.2434 & 0.0141 & 0.9253 \\
 & MLP & 0.3293 & 0.2441 & 0.0151 & 0.9257 \\
 & Random Forest & 0.3650 & 0.3684 & 0.0117 & 1.3448 \\
 & SVM & 0.3299 & 0.3108 & 0.0113 & 1.1827 \\
 & Transformer & 0.3052 & 0.2446 & 0.0112 & 0.9309 \\
 & \textbf{\textcolor{red}{CogReact}} & \textbf{\textcolor{red}{0.2703}} & \textbf{\textcolor{red}{0.2224}} & \textbf{\textcolor{red}{0.0093}} & \textbf{\textcolor{red}{0.7631}} \\
\midrule
\multirow{4}{*}{\makecell[l]{Task: Encoded String \\ Feedback: Numeric Value}}  & Decision Tree & 0.3639 & 0.3639 & 0.0112 & 1.3917 \\
 & Linear Regression & 0.3512 & 0.3469 & 0.0105 & 1.3176 \\
 & LSTM & 0.3278 & 0.2478 & 0.0142 & 0.9397 \\
 & MLP & 0.3333 & 0.2577 & 0.0145 & 0.9724 \\
 & Random Forest & 0.3600 & 0.3630 & 0.0130 & 1.3620 \\
 & SVM & 0.3245 & 0.3101 & 0.0123 & 1.1952 \\
 & Transformer & 0.3299 & 0.2481 & 0.0142 & 0.9350 \\
\bottomrule
\end{tabular}
\end{table*}

\begin{table*}
\caption{Statistical results by comparing our CogReact model in Type II (Task: Encoded String, Feedback: Video) with each of baseline model respectively in the Math Task.}
\label{table:baseline-statistics-string-math}
  \centering
\begin{tabular}{llll}
\toprule
&& \multicolumn{2}{c}{Statistical Tests for CogReact in Type II}\\ 
\cmidrule(r){3-4}
Model Input Type         & Model Type Name    & Kolmogorov-Smirnov            & Permutation  \\ 

\midrule
\multirow{4}{*}{\makecell[l]{Task: Video \\ Feedback: Video}}  & hGRU & $p\,<\,0.001$ & $p\,<\,0.001$ \\
 & LSTM + AlexNet & $p\,<\,0.001$ & $p\,<\,0.001$ \\
 & LSTM + VGG-16 & $p\,<\,0.001$ & $p\,<\,0.001$ \\
 & LSTM + ViT-B-16 & $p\,<\,0.001$ & $p\,<\,0.001$ \\
 & MLP + 3D ResNet & $p\,<\,0.001$ & $p\,<\,0.001$ \\
\midrule
\multirow{4}{*}{\makecell[l]{Task: Encoded String \\ Feedback: Video}}  & LSTM-V1 + 3D ResNet & $p\,<\,0.001$ & $p\,<\,0.001$ \\
 & LSTM-V2 + 3D ResNet & $p\,<\,0.001$ & $p\,<\,0.001$ \\
 & MLP + 3D ResNet & $p\,<\,0.001$ & $p\,<\,0.001$ \\
 & Transformer + 3D ResNet & $p\,<\,0.001$ & $p\,<\,0.001$ \\
\midrule
\multirow{4}{*}{\makecell[l]{Task: Numeric Value \\ Feedback: Video}}  & LSTM-V1 + 3D ResNet & $p\,<\,0.001$ & $p\,<\,0.001$ \\
 & LSTM-V2 + 3D ResNet & $p\,<\,0.001$ & $p\,<\,0.001$ \\
 & MLP + 3D ResNet & $p\,<\,0.001$ & $p\,<\,0.001$ \\
 & Transformer + 3D ResNet & $p\,<\,0.001$ & $p\,<\,0.001$ \\
\midrule
\multirow{4}{*}{\makecell[l]{Task: Numeric Value \\ Feedback: Numeric Value}}  & Decision Tree & $p\,<\,0.001$ & $p\,<\,0.001$ \\
 & Linear Regression & $p\,<\,0.001$ & $p\,<\,0.001$ \\
 & \textbf{\textcolor{red}{LSTM}} & \textbf{\textcolor{red}{$p\,=\,0.819$}} & \textbf{\textcolor{red}{$p\,=\,0.263$}} \\
 & MLP & $p\,<\,0.001$ & $p\,<\,0.001$ \\
 & Random Forest & $p\,<\,0.001$ & $p\,<\,0.001$ \\
 & SVM & $p\,<\,0.001$ & $p\,<\,0.001$ \\
 & \textbf{\textcolor{red}{Transformer}} & \textbf{\textcolor{red}{$p\,=\,0.920$}} & \textbf{\textcolor{red}{$p\,=\,0.332$}} \\
\midrule
\multirow{4}{*}{\makecell[l]{Task: Encoded String \\ Feedback: Numeric Value}}  & Decision Tree & $p\,<\,0.001$ & $p\,<\,0.001$ \\
 & Linear Regression & $p\,<\,0.001$ & $p\,<\,0.001$ \\
 & LSTM & $p\,<\,0.001$ & $p\,<\,0.001$ \\
 & MLP & $p\,<\,0.001$ & $p\,<\,0.001$ \\
 & Random Forest & $p\,<\,0.001$ & $p\,<\,0.001$ \\
 & SVM & $p\,<\,0.001$ & $p\,<\,0.001$ \\
 & Transformer & $p\,<\,0.001$ & $p\,<\,0.001$ \\

\bottomrule
\end{tabular}
\end{table*}

\begin{table*}
\caption{Statistical results by comparing our CogReact model in Type IV (Task: Numeric Value, Feedback: Numeric Value) with each of baseline model respectively in the Math Task.}
\label{table:baseline-statistics-numeric-math}
  \centering
\begin{tabular}{llll}
\toprule
&& \multicolumn{2}{c}{Statistical Tests for CogReact in Type IV}\\ 
\cmidrule(r){3-4}
Model Input Type         & Model Type Name    & Kolmogorov-Smirnov            & Permutation  \\ 

\midrule
\multirow{4}{*}{\makecell[l]{Task: Video \\ Feedback: Video}}  & hGRU & $p\,<\,0.001$ & $p\,<\,0.001$ \\
 & LSTM + AlexNet & $p\,<\,0.001$ & $p\,<\,0.001$ \\
 & LSTM + VGG-16 & $p\,<\,0.001$ & $p\,<\,0.001$ \\
 & LSTM + ViT-B-16 & $p\,<\,0.001$ & $p\,<\,0.001$ \\
 & MLP + 3D ResNet & $p\,<\,0.001$ & $p\,<\,0.001$ \\
\midrule
\multirow{4}{*}{\makecell[l]{Task: Encoded String \\ Feedback: Video}}  & LSTM-V1 + 3D ResNet & $p\,<\,0.001$ & $p\,<\,0.001$ \\
 & LSTM-V2 + 3D ResNet & $p\,<\,0.001$ & $p\,<\,0.001$ \\
 & MLP + 3D ResNet & $p\,<\,0.001$ & $p\,<\,0.001$ \\
 & Transformer + 3D ResNet & $p\,<\,0.001$ & $p\,<\,0.001$ \\
\midrule
\multirow{4}{*}{\makecell[l]{Task: Numeric Value \\ Feedback: Video}}  & LSTM-V1 + 3D ResNet & $p\,<\,0.001$ & $p\,<\,0.001$ \\
 & LSTM-V2 + 3D ResNet & $p\,<\,0.001$ & $p\,<\,0.001$ \\
 & MLP + 3D ResNet & $p\,<\,0.001$ & $p\,<\,0.001$ \\
 & Transformer + 3D ResNet & $p\,<\,0.001$ & $p\,<\,0.001$ \\
\midrule
\multirow{4}{*}{\makecell[l]{Task: Numeric Value \\ Feedback: Numeric Value}}  & Decision Tree & $p\,<\,0.001$ & $p\,<\,0.001$ \\
 & Linear Regression & $p\,<\,0.001$ & $p\,<\,0.001$ \\
 & \textbf{\textcolor{red}{LSTM}} & \textbf{\textcolor{red}{$p\,<\,0.001$}} & \textbf{\textcolor{red}{$p\,<\,0.001$}} \\
 & MLP & $p\,<\,0.001$ & $p\,<\,0.001$ \\
 & Random Forest & $p\,<\,0.001$ & $p\,<\,0.001$ \\
 & SVM & $p\,<\,0.001$ & $p\,<\,0.001$ \\
 & \textbf{\textcolor{red}{Transformer}} & \textbf{\textcolor{red}{$p\,<\,0.001$}} & \textbf{\textcolor{red}{$p\,<\,0.001$}} \\
\midrule
\multirow{4}{*}{\makecell[l]{Task: Encoded String \\ Feedback: Numeric Value}}  & Decision Tree & $p\,<\,0.001$ & $p\,<\,0.001$ \\
 & Linear Regression & $p\,<\,0.001$ & $p\,<\,0.001$ \\
 & LSTM & $p\,<\,0.001$ & $p\,<\,0.001$ \\
 & MLP & $p\,<\,0.001$ & $p\,<\,0.001$ \\
 & Random Forest & $p\,<\,0.001$ & $p\,<\,0.001$ \\
 & SVM & $p\,<\,0.001$ & $p\,<\,0.001$ \\
 & Transformer & $p\,<\,0.001$ & $p\,<\,0.001$ \\
 
\bottomrule
\end{tabular}
\end{table*}

\begin{table*}
\caption{Statistical results by comparing CogReact with each of baseline model and ablation model (our model variants in ablation studies) respectively in PeerEdu dataset.}
\label{table:baseline-statistics-peeredu}
  \centering
\begin{tabular}{llll}
\toprule
&& \multicolumn{2}{c}{Statistical Tests for CogReact in PeerEdu}\\ 
\cmidrule(r){3-4}
Model Input Type         & Model Type Name    & Kolmogorov-Smirnov            & Permutation  \\ 

\midrule
\multirow{4}{*}{\makecell[l]{Task: Numeric Value \\ Feedback: Numeric Value}}  & Pure DRL & $p\,<\,0.001$ & $p\,<\,0.001$ \\
 & Unencoded Hybrid DRL & $p\,=\,0.037$ & $p\,<\,0.001$ \\
 & Encoded SVM & $p\,<\,0.001$ & $p\,<\,0.001$ \\
 & LSTM & $p\,<\,0.001$ & $p\,<\,0.001$ \\
 & Transformer & $p\,<\,0.001$ & $p\,<\,0.001$ \\
 & MLP & $p\,<\,0.001$ & $p\,<\,0.001$ \\
 & SVM & $p\,<\,0.001$ & $p\,<\,0.001$ \\
 & Random Forest & $p\,<\,0.001$ & $p\,<\,0.001$ \\
 & Decision Tree & $p\,<\,0.001$ & $p\,<\,0.001$ \\
 & Linear Regression & $p\,<\,0.001$ & $p\,<\,0.001$ \\

\bottomrule
\end{tabular}
\end{table*}




\begin{table*}
\caption{Statistical results by comparing CogReact with each of baseline model and ablation model (our model variants in ablation studies) respectively in CPC dataset.}
\label{table:baseline-statistics-cpc}
  \centering
\begin{tabular}{llll}
\toprule
&& \multicolumn{2}{c}{Statistical Tests for CogReact in CPC}\\ 
\cmidrule(r){3-4}
Model Input Type         & Model Type Name    & Kolmogorov-Smirnov            & Permutation  \\ 

\midrule
\multirow{4}{*}{\makecell[l]{Task: Numeric Value \\ Feedback: Numeric Value}}  & Pure DRL & $p\,<\,0.001$ & $p\,<\,0.001$ \\
 & Unencoded Hybrid DRL & $p\,=\,0.058$ & $p\,=\,0.677$ \\
 & Encoded SVM & $p\,<\,0.001$ & $p\,<\,0.001$ \\
 & LSTM & $p\,<\,0.001$ & $p\,<\,0.001$ \\
 & Transformer & $p\,<\,0.001$ & $p\,<\,0.001$ \\
 & MLP & $p\,<\,0.001$ & $p\,<\,0.001$ \\
 & SVM & $p\,<\,0.001$ & $p\,<\,0.001$ \\
 & Random Forest & $p\,<\,0.001$ & $p\,<\,0.001$ \\
 & Decision Tree & $p\,<\,0.001$ & $p\,<\,0.001$ \\
 & Linear Regression & $p\,<\,0.001$ & $p\,<\,0.001$ \\

\bottomrule
\end{tabular}
\end{table*}



\begin{figure*}
\centering
\includegraphics[width=1\linewidth]{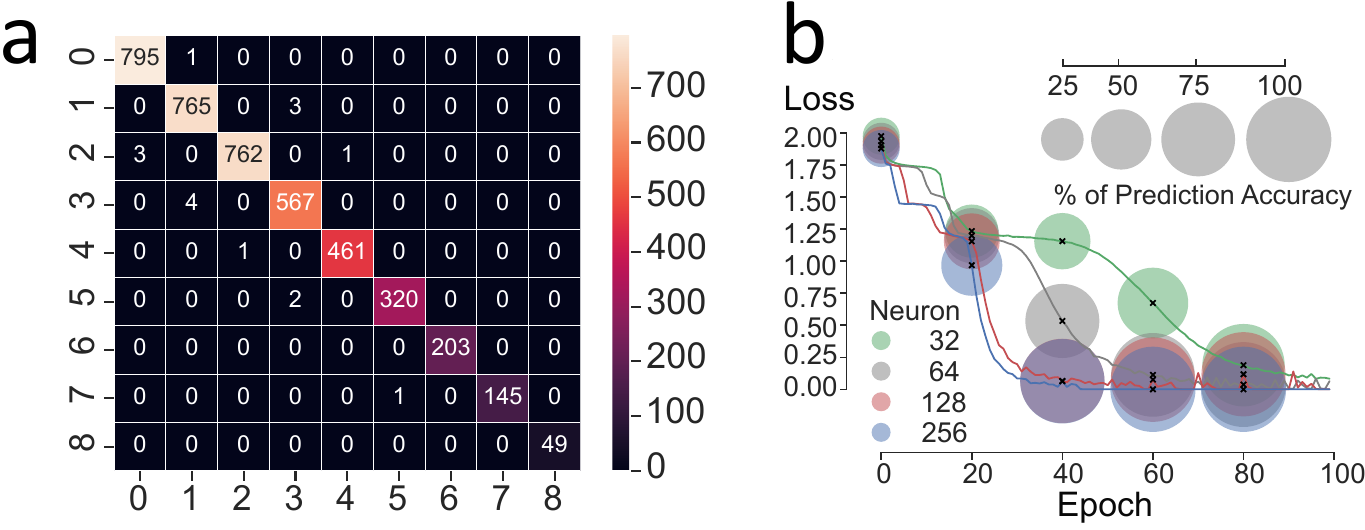}
\caption{
a. Confusion matrix (x axis: ground truth, y axis: prediction) for testing set prediction of the logical reasoning agent (LSTM neuron = 256). b. Training loss and accuracy with training epochs across four kinds of LSTM neurons of the logical reasoning agent. 
}
\label{fig:math agent result}
\end{figure*}

\begin{figure*}
\centering
\includegraphics[width=0.9\linewidth]{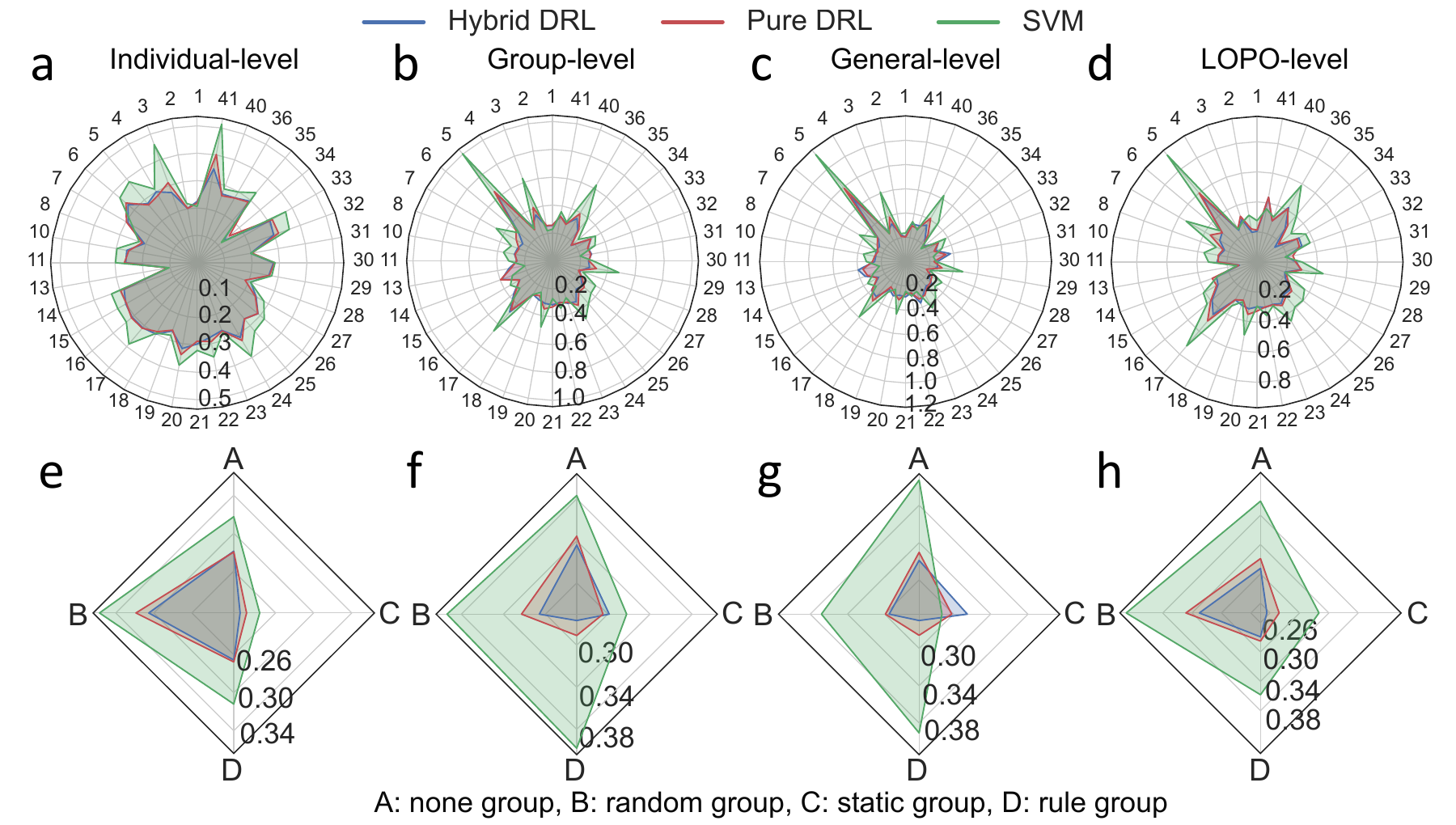}
\caption{
Evaluation results in the logical reasoning task for MAPE in different levels.
a,b,c,d,e,f,g,h: Average MAPE for each participant (a,b,c,d)/group (e,f,g,h) in predictions of testing set from Hybrid DRL agent, Pure DRL agent, and SVM model in four training strategies (a,e. Individual-Level, b,f. Group-Level, c,g. General-Level, d,h. LOPO-Level), respectively. (The number around the circle represents participant id in a,b,c,d).
}
\label{fig: comparion with pure DRL results}
\end{figure*}

\begin{figure*}
\centering
\includegraphics[width=1\linewidth]{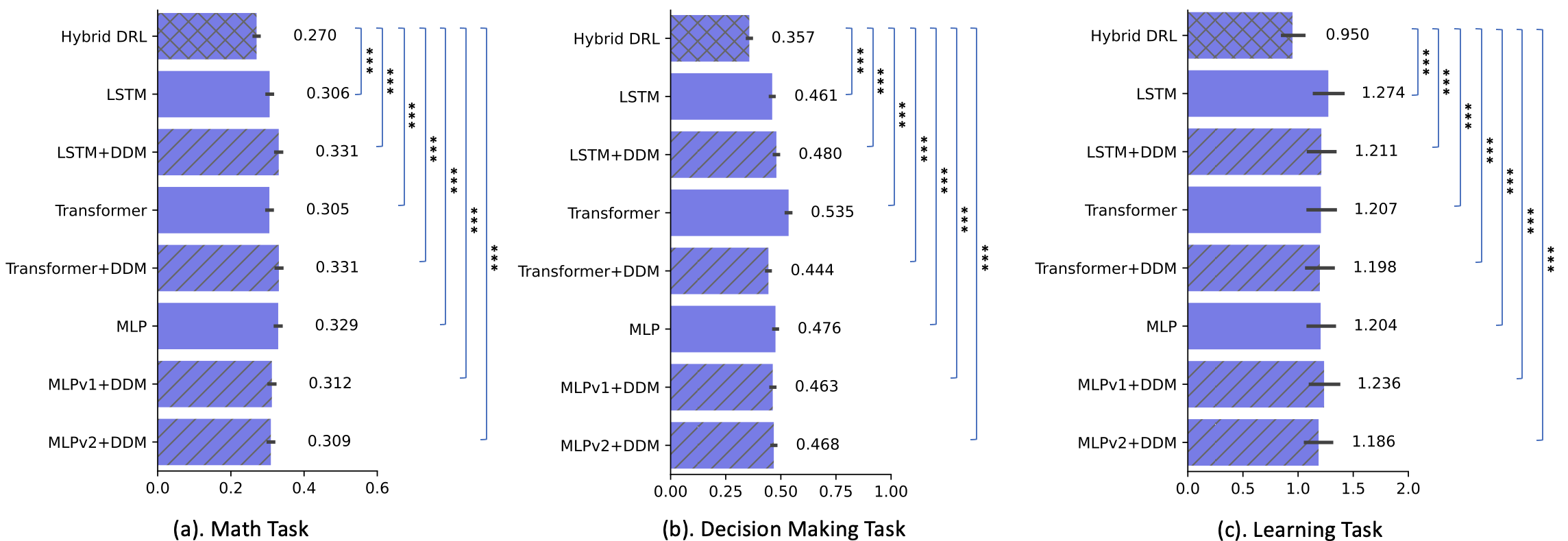}
\caption{Comparison results between our CogReact Model (Hybrid DRL) and baseline deep learning models with / without DDM in three datasets. For statistical analysis with both Kolmogorov-Smirnov test and Permutation test,  $*$ indicates $p < 0.05$, $**$ indicates $p < 0.01$, $***$ indicates $p < 0.001$. The results show that the MAPE of our model is significantly ($p < 0.001$) smaller than all deep learning models with / without DDM in all three datasets.
}
\label{fig:result-dl-ddm}
\end{figure*}

\end{document}